\PassOptionsToPackage{table}{xcolor}
\documentclass[10pt,twocolumn,letterpaper]{article}

\usepackage[pagenumbers]{cvpr} %

\usepackage{amsmath}
\usepackage{amssymb}
\usepackage{booktabs}
\usepackage{graphicx}
\usepackage{float}
\usepackage{bm}
\usepackage[table]{xcolor}

\definecolor{occrow}{HTML}{FFEEDE}
\definecolor{occgreen}{HTML}{00A651}
\definecolor{occred}{HTML}{FF0000}
\definecolor{softyellow}{RGB}{250,245,224}
\definecolor{softteal}{RGB}{228,243,237}
\definecolor{roiRed}{HTML}{D42632}
\definecolor{labelInk}{HTML}{20252B}

\newcommand{\occclass}[2]{%
  \rotatebox{90}{%
    \textcolor[HTML]{#1}{\rule{1.2ex}{1.2ex}}%
    \hspace{0.4em}#2%
  }%
}

\definecolor{cvprblue}{rgb}{0.21,0.49,0.74}
\usepackage[pagebackref,breaklinks,colorlinks,allcolors=cvprblue]{hyperref}

\def\confName{CVPR}
\def\confYear{2027}

\title{CGOcc: Learning Cross-Gaussian Dependencies for 3D Semantic Occupancy Prediction}

\author{%
Zirui HUANG$^{1,}$\thanks{Zirui HUANG and Jiongle He contributed equally and are co-first authors.}\quad
Jiongle He$^{1,*}$\quad
Yunfan Zeng$^{1}$\quad
Yuanzhi Lu$^{2}$\quad
Guangye Li$^{3}$\\
Boyan Tang$^{1}$\quad
Weite SU$^{4}$\quad
Pedro V. Sander$^{1,}$\thanks{Corresponding authors: Pedro V. Sander (\texttt{psander@ust.hk}) and Amy Hui Zhang (\texttt{amyzhang@bnbu.edu.cn}).}\quad
Amy Hui Zhang$^{5,\dagger}$\\[3pt]
{\small $^{1}$Hong Kong University of Science and Technology\quad $^{2}$Nanyang Technological University}\\
{\small $^{3}$The Hong Kong University of Science and Technology (Guangzhou)}\\
{\small $^{4}$The University of Hong Kong\quad $^{5}$Beijing Normal-Hong Kong Baptist University}%
}

\begin{document}

\maketitle

\begin{abstract}
\textit{
In Gaussian-based semantic occupancy prediction, overlapping primitives
can provide ambiguous class predictions, making their joint interpretation
important for voxel classification.
However, existing methods using spatially weighted mixtures neglected explicit modeling of cross-primitive dependencies, limiting the readout's ability to utilize category-specific relational evidence.
Hence, we introduce \textbf{CGOcc}, the first supervised multi-camera framework that explicitly incorporates class-conditional second-order \textbf{C}ross-\textbf{G}aussian interactions into Gaussian-to-voxel semantic readout for 3D semantic \textbf{Occ}upancy prediction.
CGOcc integrates appearance and geometric observations to refine Gaussian states and encodes cross-Gaussian dependencies in a diagonal-plus-low-rank representation.
To incorporate these dependencies into voxel classification, CGOcc utilizes a class-conditional projection to extract diagonal and signed cross-Gaussian statistics without constructing a dense dependency matrix. The readout then refines occupied-class probabilities using these statistics while preserving a learned reference’s occupied mass and free probability.
Extensive experiments on nuScenes demonstrate CGOcc achieves state-of-the-art performance, with larger gains at high Gaussian overlap and improved predictions near boundaries.
Ablation studies validate the effectiveness of covariance-aware prediction and support its synergy with geometric interaction.
}
\end{abstract}

\begin{figure}[!t]
    \centering
    \includegraphics[width=\columnwidth]{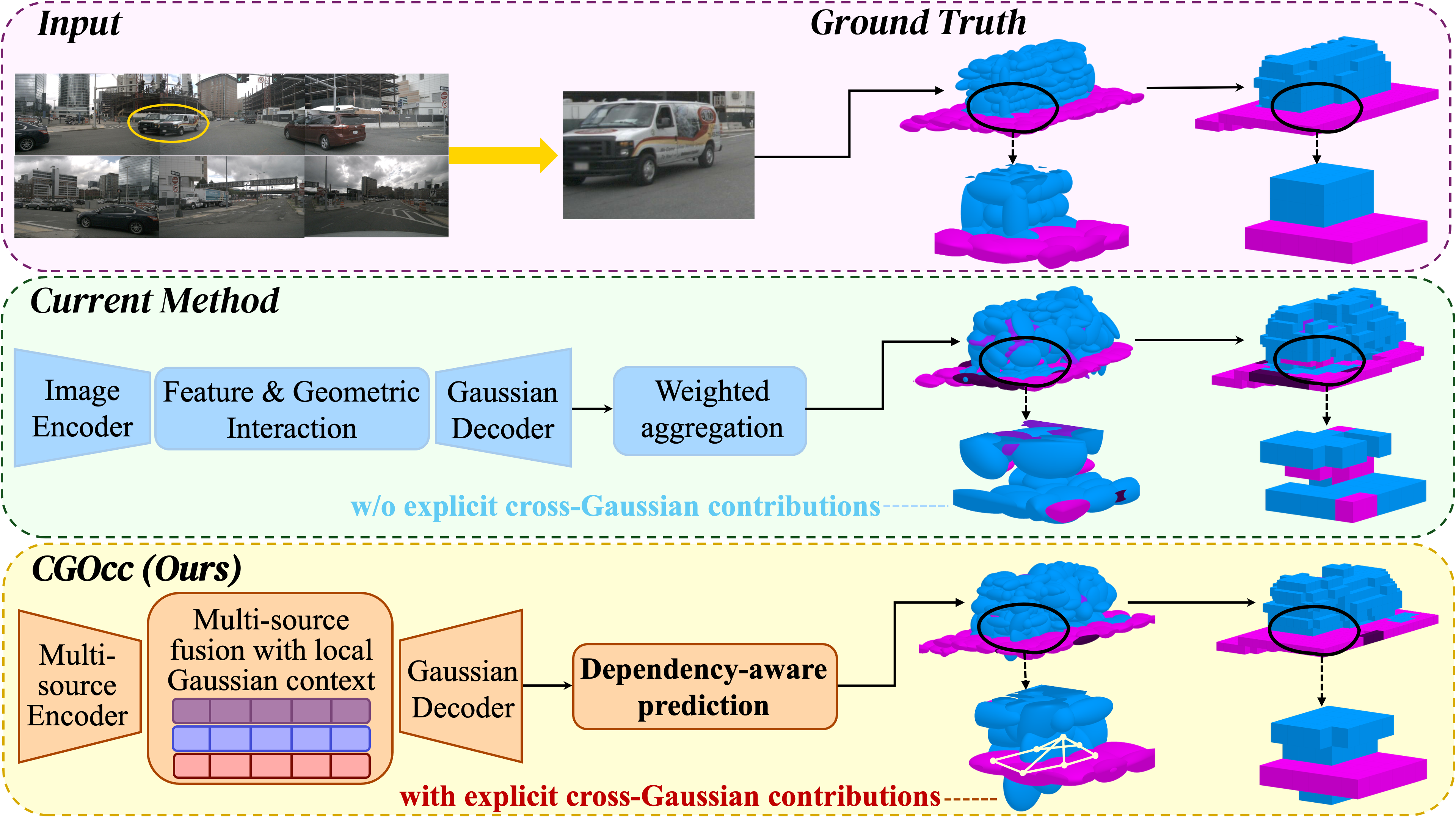}

    \caption{
        \textbf{Cross-Gaussian dependencies for semantic occupancy.}
        A schematic vehicle–road example compares weighted aggregation (middle) with CGOcc (bottom), with input and ground truth above. Pale yellow edges indicate selected cross-Gaussian dependencies.
    }
    \label{fig:teaser}
\end{figure}

\section{Introduction}
\label{sec:introduction}

Camera-based 3D semantic occupancy prediction describes scene geometry
and semantics beyond bounding boxes for autonomous
perception~\cite{tian2023occ3d,wei2023surroundocc}.
Alongside dense volumetric and tri-perspective
representations~\cite{wei2023surroundocc,huang2023tpvformer}, semantic
Gaussians offer adaptive, continuous scene
representations~\cite{huang2024gaussianformer,huang2025gaussianformer2}.
Their overlapping supports make voxel classification depend on
the joint interpretation of potentially competing primitive predictions.

We identify a structural limitation of mixture-based semantic readout.
GaussianFormer-2 combines individual Gaussian class probabilities
through a spatially weighted mixture~\cite{huang2025gaussianformer2}.
Although decoder interactions can influence these individual
predictions, the mixture contains no explicit class-conditional
cross-Gaussian interaction term.
At fixed spatial weights and primitive class probabilities, different
cross-Gaussian dependency structures therefore produce the same
semantic mixture.
The readout cannot directly use these differences to distinguish
ambiguous configurations of overlapping primitives.
This limitation motivates explicit relational evidence at the point
where Gaussian predictions are converted into voxel semantics,
as illustrated in Fig.~\ref{fig:teaser}.

To address this limitation, we introduce \textbf{CGOcc}, the first
supervised multi-camera framework to explicitly incorporate
class-conditional second-order \textbf{C}ross-\textbf{G}aussian
interactions into Gaussian-to-voxel semantic readout for 3D semantic
\textbf{Occ}upancy prediction.
CGOcc learns a diagonal-plus-low-rank dependency
kernel~\cite{monteiro2020ssn}, distinct from each primitive's spatial
covariance.
Factorized projection through spatial weights and class probabilities
extracts diagonal and signed cross-Gaussian statistics without
enumerating pairs or forming a dense
matrix~\cite{rendle2010fm,he2017nfm}.
A learned reference combines voxel context and diagonal statistics.
Cross statistics drive bounded corrections that reweight its
occupied-class probabilities while preserving its occupied mass
and free-space probability, and exactly recovers the reference when
the cross input vanishes.

Learning these dependencies also requires Gaussian states that
integrate complementary observations.
DINOv2 provides visual features~\cite{oquab2024dinov2},
Metric3Dv2 provides metric depth~\cite{hu2024metric3dv2},
and VGGT provides multi-view geometry~\cite{wang2025vggt}.
GaussTR connects foundation-model outputs to Gaussian representations
through rendered feature and depth
alignment~\cite{jiang2025gausstr}.
In CGOcc, we integrate appearance and geometric observations around
evolving Gaussians through a geometry-conditioned multi-source
interaction module.
Because the usefulness of observations varies across views and
locations, the module combines cross-source observation exchange
with geometry-biased query attention.
Local Gaussian-neighbor messages further incorporate spatial context.
The resulting Gaussian states and observations provide the inputs
for dependency modeling, connecting observation integration to
relational semantic readout.

Extensive experiments on nuScenes show that CGOcc achieves state-of-the-art
performance among the compared methods.
In spatial analyses, mIoU gains increase across high-overlap bins and peak
in the highest, and CGOcc improves
semantic predictions near boundaries, with the largest margin occurring
at 0.4\,m from ground-truth interfaces.
The full class-conditional readout outperforms its
ablated variants, while paired ablations support conditional synergy between decoder
mechanisms and the covariance-aware readout.

In summary, our main contributions are as follows:
\begin{itemize}
    \item CGOcc introduces the first class-conditional second-order semantic readout to explicitly incorporate signed cross-Gaussian interactions into voxel classification, while preserving the learned reference’s occupied mass and free-space probability.

    \item The geometry-conditioned multi-source interaction
    module integrates heterogeneous observations around evolving
    Gaussian states and propagates local Gaussian context to support
    refinement and dependency modeling.

    \item CGOcc achieves state-of-the-art performance on nuScenes,
    with particularly pronounced gains in regions of high Gaussian
    overlap and more accurate semantic predictions near boundaries.
\end{itemize}

\section{Related work}
\label{sec:related_work}

\textbf{Gaussian Representations for 3D Semantic Occupancy.}
The transition from scene-specific Gaussian rendering~\cite{kerbl2023gaussians}
to semantic voxel occupancy~\cite{huang2024gaussianformer} connects
representation design with rendering-based supervision.
On the representation side, methods separate occupancy from semantic
mixtures~\cite{huang2025gaussianformer2}, combine depth-guided initialization
with decoupled geometric and semantic aggregation~\cite{qian2026splatssc},
and incorporate scene evolution, spatial-temporal fusion, and incremental
memory~\cite{zuo2025gaussianworld,yan2026stgs,wu2025embodiedocc}.
Rendering provides self- or weak
supervision~\cite{gan2025gaussianocc,boeder2025gaussianflowocc,sun2026gsrender},
including GaussTR's foundation-model alignment of rendered features and
depth for open-vocabulary inference~\cite{jiang2025gausstr}.
GaussRender extends this supervisory role to architectures without
Gaussian internal representations through projective
supervision~\cite{chambon2025gaussrender}.
Here, spatial covariance describes individual Gaussian shape, leaving
cross-primitive semantic dependencies as a separate modeling question.

\textbf{Foundation Model Priors and Gaussian Feature Interaction.}
DINOv2, Metric3Dv2, and VGGT supply complementary visual features,
metric depth and normals, and multi-view
geometry~\cite{oquab2024dinov2,hu2024metric3dv2,wang2025vggt}.
While Feature 3DGS and LangSplat transfer visual and language features
into scene-level representations~\cite{zhou2024feature3dgs,qin2024langsplat},
feed-forward occupancy methods adapt frozen geometric features and
extend surface priors into volumetric support~\cite{yan2026vg3s,zhou2026gpocc},
with further integration of ray geometry and image
context~\cite{zhou2026gpoccplus,liu2026occanyscene}.
Their integration depends on how observations interact and contribute:
Set Transformer models observation interactions~\cite{lee2019settransformer},
deformable attention localizes sampling~\cite{zhu2021deformabledetr},
and VALA uses rendering visibility for language aggregation~\cite{wang2026vala}.
Our decoder combines appearance and geometric priors with multi-view
evidence and local Gaussian context.

\begin{figure*}[!t]
    \centering
    \includegraphics[width=\textwidth]{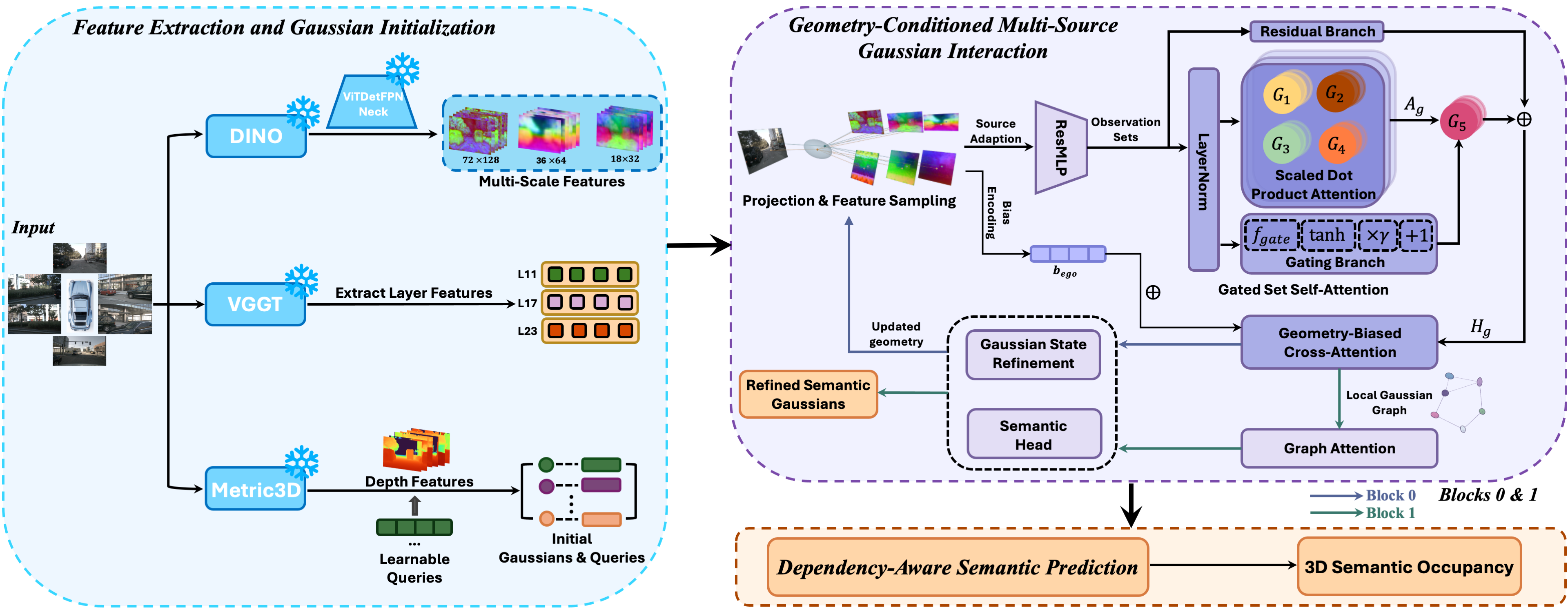}
    \caption[Overview of CGOcc]{
    \textbf{Overview of CGOcc.}
    Depth-initialized Gaussians undergo two geometry-conditioned
    multi-source interaction blocks before dependency-aware semantic
    readout (Fig.~\ref{fig:framework2}).
    Only Block 1 applies local graph attention; snowflakes denote frozen modules.
    }
    \label{fig:overall_framework}
\end{figure*}

\textbf{Structured Dependencies and Covariance-Aware Semantic Aggregation.}
Structured prediction captures pairwise label
compatibility~\cite{krahenbuhl2011densecrf,zheng2015crfrnn}, spatially
correlated aleatoric uncertainty through Gaussian logits with
diagonal-plus-low-rank covariance~\cite{monteiro2020ssn}, and segmentation
ambiguity or correlated annotation
errors~\cite{kohl2018probunet,ribeiro2025flowssn,neverova2019correlated},
whereas factorization and bilinear models learn predictive feature
interactions~\cite{rendle2010fm,he2017nfm,xiao2017afm,lin2015bilinear,gao2016compactbilinear}.
For semantic mapping, Bayesian kernel inference and ConvBKI fuse spatial
evidence~\cite{gan2020semanticbki,wilson2024convbki}, with E2-BKI adding
evidential uncertainty~\cite{kim2026e2bki}.
Continuous Semantic Splatting maintains per-ellipsoid Dirichlet semantic
beliefs, but computes rendered variance using squared weights under
independence, excluding cross-ellipsoid covariance~\cite{wilson2025css}.
Motivated by this distinction, our method projects learned
cross-Gaussian dependencies through spatial support and class
probabilities, using the resulting class-conditional statistics
as discriminative signals to refine voxel-level semantic predictions.

\section{Methodology}

This section presents a comprehensive elaboration of the
CGOcc framework.
We initialize Gaussian centers from depth
(Sec.~\ref{sec:problem_formulation}) and refine their states using
multi-view appearance and geometric features
(Sec.~\ref{sec:gaussian_interaction}).
We then model dependencies between Gaussians and project them into
class-specific statistics for voxel semantic prediction
(Sec.~\ref{sec:dependency_readout}).
Voxel and auxiliary Gaussian supervision jointly train the task-specific
modules (Sec.~\ref{sec:training_objective}).
An overview of the CGOcc framework is depicted in Fig.~\ref{fig:overall_framework}.

\subsection{Problem Formulation and Gaussian Representation}
\label{sec:problem_formulation}

Given surround-view RGB images
$\{I_m,\mathbf K_m,\mathbf T_m\}_{m=1}^{M}$ with intrinsics
$\mathbf K_m$ and camera-to-ego extrinsics $\mathbf T_m$, we predict voxel distributions
$\mathbf P^{X\times Y\times Z\times(C+1)}$ over $C$ occupied
classes and free space, $\mathcal C=\mathcal C_{\mathrm{occ}}\cup\{\mathrm{free}\}$.
Voxel labels supervise $\mathbf P$ with predicted classes
$\hat Y_v=\arg\max_{c\in\mathcal C}P_{v,c}$ at ego-frame voxel centers
$\mathbf x_v$, $v\in\mathcal V$.

We represent the scene with $N$ anisotropic
Gaussians~\cite{huang2024gaussianformer}, each with center
$\boldsymbol\mu_g\in\mathbb R^3$, opacity $o_g\in[0,1]$,
query $\mathbf q_g\in\mathbb R^d$, and spatial covariance
$\boldsymbol\Sigma_g=\mathbf R_g
\operatorname{diag}(\mathbf s_g^2)\mathbf R_g^\top$,
where $\mathbf s_g$ contains the scales and $\mathbf R_g$
is the rotation of a unit quaternion.
A shared head predicts occupied-class probabilities
$\mathbf p_g=\operatorname{softmax}(f_{\mathrm{sem}}(\mathbf q_g))$,
while free space is inferred at the voxel level.
Each Gaussian
contributes within its splatting neighborhood as:

\begin{equation}
    \begin{aligned}
        \mathbf d_{vg} &= \mathbf x_v - \boldsymbol\mu_g,
        \quad
        \rho_{vg} = o_g \exp\!\left(
            -\tfrac{1}{2}\mathbf d_{vg}^{\top}
            \boldsymbol\Sigma_g^{-1}\mathbf d_{vg}
        \right).
    \end{aligned}
    \label{eq:gaussian_spatial_contribution}
\end{equation}

Frozen DINOv2~\cite{oquab2024dinov2} with GaussTR's frozen pyramid
neck~\cite{jiang2025gausstr} provides appearance features.
Frozen VGGT jointly extracts geometric features from the views and
we retain outputs at zero-based block indices $\{11,17,23\}$,
corresponding to the three deeper feature taps used by its
DPT head~\cite{wang2025vggt}.
Following GaussTR, we initialize Gaussian centers by calibrated
back-projection of Metric3D v2 depth~\cite{hu2024metric3dv2}
at fixed image references.

\subsection{Geometry-Conditioned Multi-Source Gaussian Interaction}
\label{sec:gaussian_interaction}

Appearance and geometric observations provide complementary cues for Gaussian refinement, but their usefulness can vary across views and locations. CGOcc therefore allows these observations to exchange information and exploits camera-relative geometry to guide their aggregation into each Gaussian query. Refinement proceeds through two successive blocks, with neighbor messages added in the second to complement each Gaussian’s sampled observations with local spatial context.

\begin{figure*}[!t]
    \centering
    \includegraphics[width=\textwidth]{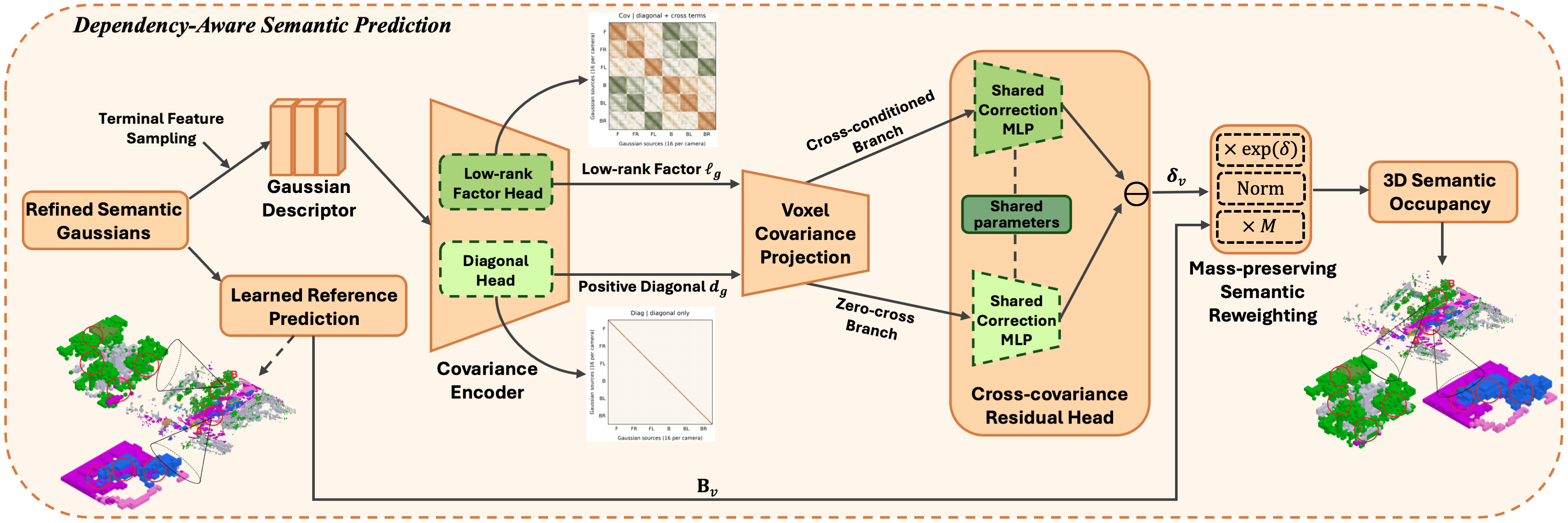}
    \caption{
    \textbf{Overview of Dependency-Aware Semantic Prediction.}
    Class-conditional projection extracts diagonal and cross-Gaussian
    statistics from a diagonal-plus-low-rank representation.
    Shared-head differencing guides semantic reweighting of the learned
    reference, preserving its occupied mass and free probability.
    }
    \label{fig:framework2}
\end{figure*}

\textbf{Local observation construction.}
In block $\ell\in\{0,1\}$, we sample features at
$\mathbf c_g^{(\ell)}=\boldsymbol\mu_g^{(\ell)}
+\boldsymbol\delta_{g,\mathrm{read}}^{(\ell)}$
and $\mathbf c_g^{(\ell)}+\Delta\mathbf x_g^{(\ell)}$,
where the additional offset is predicted from the query and scaled
by the current Gaussian size~\cite{zhu2021deformabledetr}.
Calibrated bilinear sampling and source-specific residual
adapters~\cite{chen2022adaptformer} map valid image features into
a shared $d$-dimensional space.
Appearance levels are fused with learned masked-softmax weights,
while three VGGT levels remain separate.
Together with camera, source, and sampling embeddings, these features
form the observation set $\mathbf O_g^{(\ell)}\in\mathbb R^{S\times d}$
with valid indices $\mathcal J_g^{(\ell)}$ and $S=2M(1+3)$.

\textbf{Geometry-conditioned observation interaction.}
We apply masked self-attention across
observations~\cite{lee2019settransformer} and, inspired by attention-output gating~\cite{qiu2025gatedattention}, use observation-conditioned
residual gating to modulate the resulting updates before query aggregation.
Omitting block indices, we denote the projected self-attention output from \(\operatorname{LN}(\mathbf O_g)\) by \(\mathbf A_g\) and write the gated residual update as
\begin{equation}
    \mathbf H_g
    =\mathbf O_g+
    \bigl[\mathbf1+\gamma\tanh(f_{\mathrm{gate}}
    (\operatorname{LN}(\mathbf O_g)))\bigr]\odot\mathbf A_g,
    \label{eq:gated_observation_interaction}
\end{equation}

\noindent Where $f_{\mathrm{gate}}$ maps normalized observation features to
gating signals and $\gamma$ is a learnable gating scale.
The query then attends to {\small $\mathbf H_g$} using geometry-biased
cross-attention~\cite{ying2021graphormer}.
A learned bounded mapping of camera-relative geometry, source
membership, and sampling metadata supplies the head-specific bias
$b_{gj}^{h}$.
With head-specific projections of normalized {\small $\mathbf q_g$} and {\small $\mathbf H_g$}, the update is
{
\small
\begin{equation}
    \begin{aligned}
        a_{gj}^{h}
        &=\operatorname{softmax}_{j\in\mathcal J_g}
        \!\left(
        \frac{\langle\mathbf q_g^h,\mathbf H_{gj}^{h,\mathrm{k}}\rangle}
        {\sqrt{d_h}}+b_{gj}^{h}\right),\\
        \mathbf u_g
        &=\mathbf q_g+W_{\mathrm{out}}
        \operatorname{Concat}_{h}\!\left(
        \sum_{j\in\mathcal J_g}a_{gj}^{h}
        \mathbf H_{gj}^{h,\mathrm{v}}\right).
    \end{aligned}
    \label{eq:geometry_conditioned_logits}
\end{equation}
}

\noindent Where $d_h$ is the head dimension.
A feed-forward residual produces $\widetilde{\mathbf q}_g^{(\ell)}$;
all attention aggregates valid observations only.

\textbf{Gaussian interaction and refinement.}
We defer graph attention~\cite{velickovic2018gat,zhao2021pointtransformer}
to the second block so that local message passing builds on an initial
round of Gaussian refinement from multi-view observations.
The first block sets
$\mathbf q_g^{(1)}=\widetilde{\mathbf q}_g^{(0)}$,
whereas the second applies local graph
attention to obtain $\mathbf q_g^{(2)}$.
Each Gaussian receives messages from up to $K_{nbr}$ nearest initially valid
non-self neighbors within radius $R$, with relative positions
conditioning attention and positional messages.
After each block, shared heads decode class probabilities and geometric
updates from the resulting queries.
Cumulative center displacement is bounded relative to initialization,
and the updated geometry guides feature sampling in the next block.
The final states are used for semantic readout
(Sec.~\ref{sec:dependency_readout}), while
intermediate predictions provide auxiliary supervision without feeding
back into subsequent queries.

\subsection{Dependency-Aware Semantic Prediction}
\label{sec:dependency_readout}

At voxel readout, however, a spatially weighted mixture of these predictions contains no explicit cross-Gaussian dependency term. CGOcc therefore projects learned dependencies through spatial weights and class probabilities to obtain class-conditional voxel statistics. These statistics guide bounded semantic corrections to a learned reference while preserving its occupied mass and free probability (Fig.~\ref{fig:framework2}).

\textbf{Initial Gaussian-to-voxel aggregation.} Following Gaussian-based
readout~\cite{jiang2025gausstr,yan2026vg3s,kerbl2023gaussians},
we truncate Eq.~\eqref{eq:gaussian_spatial_contribution} to finite
splatting neighborhoods, obtaining contributions $\omega_{vg}$.
Their sum gives density $S_v$; normalization by $S_v$ gives
weights $w_{vg}$, set to zero at unsupported voxels.
Utilizing the occupancy--semantics decomposition in
GaussianFormer-2~\cite{huang2025gaussianformer2}, we compute
{
\small
\begin{equation}
    \begin{aligned}
        \bar p^{(0)}_{vc}
        =\frac{\sum_g\omega_{vg}p_{gc}+\alpha\pi_c}{S_v+\alpha},
        \qquad
        \mathbf P_v^{(0)}=[\rho_v\bar{\mathbf p}_v^{(0)},\,1-\rho_v].
    \end{aligned}
\end{equation}
}

\noindent The learned prior $\boldsymbol\pi$ smooths semantics with strength
$\alpha>0$, including unsupported voxels.
A sigmoid of learned log-density calibration plus a depth-derived bias
gives occupied probability $\rho_v$.

\textbf{Cross-Gaussian dependency modeling.}
A shared encoder maps Gaussian geometry, opacity, semantics, and a
terminal observation summary to $\mathbf h_g$.
Heads parameterize a diagonal-plus-low-rank kernel~\cite{monteiro2020ssn}:
{
\small
\begin{equation}
    \begin{gathered}
        \boldsymbol\ell_g=\tanh(f_\ell(\mathbf h_g)),
        \qquad
        d_g=\operatorname{softplus}(f_d(\mathbf h_g))+d_{\min},\\
        K=\operatorname{diag}(\mathbf d)+LL^\top.
    \end{gathered}
    \label{eq:gaussian_dependency_kernel}
\end{equation}
}

\noindent Here $L_{g,:}=\boldsymbol\ell_g^\top$, $\boldsymbol\ell_g\in\mathbb R^r$, and $d_{\min}>0$.
The positive-semidefinite
$K\in\mathbb R^{N\times N}$ models task-learned dependencies,
not spatial shape or calibrated error covariance.

\textbf{Class-conditional statistical projection.}
With $a_{vgc}=w_{vg}p_{gc}$ and
$\mathbf a_{vc}=(a_{v1c},\ldots,a_{vNc})^\top$, projection separates
individual and signed cross-Gaussian contributions as $\mathbf a_{vc}^{\top}K\mathbf a_{vc}=D_{vc}+X_{vc}$, where:
{
\small
\begin{equation}
    \begin{aligned}
        D_{vc}&=\sum_g a_{vgc}^{2}
        \bigl(d_g+\|\boldsymbol\ell_g\|_2^2\bigr),\\
        X_{vc}&=\Bigl\|\sum_g a_{vgc}\boldsymbol\ell_g\Bigr\|_2^2
        -\sum_g a_{vgc}^{2}\|\boldsymbol\ell_g\|_2^2.
    \end{aligned}
    \label{eq:class_conditional_projection}
\end{equation}
}

\noindent Factorization~\cite{rendle2010fm,he2017nfm} avoids dense $K$ and pair enumeration.
For $\epsilon_X>0$, readout uses normalized descriptors:
{
\small
\begin{equation}
    m_{vc}=\frac{D_{vc}}{1+D_{vc}},\qquad
    \xi_{vc}=\frac{X_{vc}}{D_{vc}+\epsilon_X+|X_{vc}|}.
\end{equation}
}
Context $\mathbf z_v$ uses spatially weighted
pooling~\cite{zaheer2017deepsets}.
We also define the effective support as
$n_v^{\mathrm{eff}}=(\sum_g w_{vg}^{2})^{-1}$.
For unsupported voxels, pooled statistics and effective support
are set to zero.

\textbf{Mass-preserving semantic correction.}
The reference $\mathbf B_v=F_B(\mathbf P_v^{(0)},\mathbf z_v^B,\mathbf m_v^B)$
uses separate encoder and factor heads (superscript $B$).
Its context $\mathbf z_v^B$ and diagonal descriptor $\mathbf m_v^B$
produce bounded semantic and occupancy adjustments $\mathbf s_v^B$
and $b_v^B$.
Without explicit cross input, it produces
{
\small
\begin{equation}
    \begin{aligned}
        \widehat\rho_v
        &=\frac{\rho_v e^{b_v^B}}{\rho_v e^{b_v^B}+1-\rho_v},\\
        \mathbf B_v
        &=\bigl[\widehat\rho_v\operatorname{softmax}
        (\log\bar{\mathbf p}_v^{(0)}+\mathbf s_v^B),\,
        1-\widehat\rho_v\bigr].
    \end{aligned}
    \label{eq:learned_reference_prediction}
\end{equation}
}
Writing
$H_v(\boldsymbol\xi)=H_\psi(\mathbf z_v,\mathbf m_v,\boldsymbol\xi)$,
we compute:
{\small
\begin{equation}
    \mathbf r_v=H_v(\boldsymbol\xi_v)-H_v(\mathbf0),
    \qquad
    \boldsymbol\delta_v
    =\frac{\tau}{2}\operatorname{ctr}\!\left[
    \tanh\!\left(\frac{2\mathbf r_v}{\tau}\right)\right].
    \label{eq:conditional_cross_correction}
\end{equation}
}

\noindent Both evaluations share parameters, context, and diagonal statistics;
$H_v(\mathbf0)$ centers the correction and is distinct from $\mathbf B_v$.
Here $\operatorname{ctr}$ subtracts the occupied-class mean, and
$\tau>0$ bounds correction magnitudes componentwise.
With reference occupied mass $M_v=\sum_{k\in\mathcal C_{\mathrm{occ}}}B_{vk}>0$,
normalized exponential reweighting~\cite{kivinen1997exponentiated} applies to
$c\in\mathcal C_{\mathrm{occ}}$ when $S_v>0$ and $n_v^{\mathrm{eff}}>1$
(up to numerical tolerances):

{
\small
\begin{equation}
    P_{vc}
    =M_v\frac{B_{vc}e^{\delta_{vc}}}
    {\sum_{k\in\mathcal C_{\mathrm{occ}}}
    B_{vk}e^{\delta_{vk}}},
    \qquad
    P_{v,\mathrm{free}}
    =B_{v,\mathrm{free}}.
    \label{eq:mass_preserving_prediction}
\end{equation}
}

\noindent Otherwise, we retain $\mathbf B_v$.
The update preserves occupied mass and free probability;
zero cross input gives zero correction and exactly recovers $\mathbf B_v$.

\subsection{Training Objective}
\label{sec:training_objective}

With feature extractors frozen, voxel and auxiliary Gaussian losses jointly supervise final predictions and primitive states before aggregation.
Final voxel predictions receive class-weighted negative log-likelihood
$\mathcal L_{\mathrm{NLL}}$ and Lov\'asz--Softmax
loss $\mathcal L_{\mathrm{Lov}}$~\cite{berman2018lovasz}
on camera-visible voxels, with the latter averaged over all occupied
classes.
At terminal Gaussian centers, auxiliary supervision combines
class-weighted semantic cross-entropy and occupancy binary
cross-entropy as $\mathcal L_{\mathrm{G\text{-}aux}}
    =\mathcal L_{\mathrm{G\text{-}sem}}^{\mathrm{CE}}
    +\lambda_{\mathrm{occ}}\mathcal L_{\mathrm{G\text{-}occ}}^{\mathrm{BCE}}.$
It uses initially valid, in-volume Gaussians without visibility
filtering; its semantic term excludes free-space targets.

We also apply class-weighted semantic cross-entropy
$\mathcal L_{\mathrm{mid}}$ to the first block for deep
supervision~\cite{lee2015deeplysupervised}.
Targets are sampled at detached updated centers and restricted to
initially valid, in-volume Gaussians with camera-visible occupied
labels, so target assignment provides no gradient to the centers.
The overall objective is
\begin{equation}
    \begin{aligned}
        \mathcal L
        ={}&\mathcal L_{\mathrm{NLL}}
        +\lambda_{\mathrm{Lov}}\mathcal L_{\mathrm{Lov}}
        +\lambda_{\mathrm{aux}}\mathcal L_{\mathrm{G\text{-}aux}}
        +\lambda_{\mathrm{mid}}\mathcal L_{\mathrm{mid}},
    \end{aligned}
    \label{eq:training_objective}
\end{equation}
where the $\lambda$ coefficients balance the loss terms.

\begin{table*}[!t]
\centering

\caption{\textbf{3D semantic occupancy prediction results on the
Occ3D-nuScenes validation set.}
$^{\dagger}$GaussianFormer-2 uses the 128-channel configuration
($\mathrm{Ch.}=128$).
$^{\ddagger}$RenderOcc denotes the variant trained with both
3D voxel labels and LiDAR-derived 2D semantic and depth labels.
Column-wise best results are \textbf{bold} and second best results
are \underline{underlined}.}
\label{tab:gaussrender_occ3d}

\small
\setlength{\tabcolsep}{3pt}
\renewcommand{\arraystretch}{1.08}
\resizebox{\textwidth}{!}{%
\begin{tabular}{l|cc|*{17}{c}}
\toprule
Method & OccIoU $\uparrow$ & mIoU $\uparrow$
& \occclass{8C8C8C}{others}
& \occclass{FF7832}{barrier}
& \occclass{FFB4C8}{bicycle}
& \occclass{FFFF00}{bus}
& \occclass{0096FF}{car}
& \occclass{00FFFF}{const. veh.}
& \occclass{B4B400}{motorcycle}
& \occclass{FF0000}{pedestrian}
& \occclass{FFF096}{traffic cone}
& \occclass{964600}{trailer}
& \occclass{A032FF}{truck}
& \occclass{FF00FF}{drive. surf.}
& \occclass{8C8C8C}{other flat}
& \occclass{4B004B}{sidewalk}
& \occclass{96F03C}{terrain}
& \occclass{E6E6FF}{manmade}
& \occclass{00AF00}{vegetation} \\
\midrule

MonoScene~\cite{cao2022monoscene}
& 29.55 & 6.06
& 1.75 & 7.23 & 4.26 & 4.93 & 9.38 & 5.67 & 3.98 & 3.01 & 5.90
& 4.45 & 7.17 & 14.91 & 6.32 & 7.92 & 7.43 & 1.01 & 7.65 \\

BEVDet~\cite{huang2021bevdet}
& 38.23 & 19.38
& 4.39 & 30.31 & 0.23 & 32.26 & 34.47 & 12.97 & 10.34 & 10.36 & 6.26
& 8.93 & 23.65 & 52.27 & 24.61 & 26.06 & 22.31 & 15.04 & 15.10 \\

OccFormer~\cite{zhang2023occformer}
& 41.95 & 21.93
& 5.94 & 30.29 & 12.32 & 34.40 & 39.17 & 14.44 & 16.45 & 17.22 & 9.27
& 13.90 & 26.36 & 50.99 & 30.96 & 34.66 & 22.73 & 6.76 & 6.97 \\

BEVStereo~\cite{li2023bevstereo}
& 44.33 & 24.51
& 5.73 & 38.41 & 7.88 & 38.70 & 41.20 & 17.56 & 17.33 & 14.69 & 10.31
& 16.84 & 29.62 & 54.08 & 28.92 & 32.68 & 26.54 & 18.74 & 17.49 \\

BEVFormer~\cite{li2022bevformer}
& 50.78 & 26.88
& 5.85 & 37.83 & 17.87 & \underline{40.44} & 42.43 & 7.36 & 23.88
& 21.81 & \underline{20.98}
& 22.38 & 30.70 & 55.35 & 28.36 & 36.0 & 28.06 & 20.04 & 17.69 \\

CTF-Occ~\cite{tian2023occ3d}
& 51.24 & 28.53
& 8.09 & 39.33 & 20.56 & 38.29 & 42.24 & 16.93 & 24.52
& \textbf{22.72} & \textbf{21.05}
& \underline{22.98} & 31.11 & 53.33 & 33.84 & 37.98 & 33.23 & 20.79 & 18.0 \\

RenderOcc$^{\ddagger}$~\cite{pan2024renderocc}
& 48.25 & 26.11
& 4.84 & 31.72 & 10.72 & 27.67 & 26.45 & 13.87 & 18.2 & 17.67 & 17.84
& 21.19 & 23.25 & \underline{63.2}
& \underline{36.42} & \textbf{46.21}
& \underline{44.26} & 19.58 & 20.72 \\

TPVFormer~\cite{huang2023tpvformer}
& 50.82 & 27.83
& 7.22 & 38.90 & 13.67 & \textbf{40.78} & 45.90 & 17.23 & 19.99 & 18.85 & 14.30
& \textbf{26.69} & \textbf{34.17}
& 55.65 & 35.47 & 37.55 & 30.70 & 19.40 & 16.78 \\

SurroundOcc~\cite{wei2023surroundocc}
& 53.13 & 29.21
& 8.64 & \underline{40.12} & \underline{23.36}
& 39.89 & 45.23 & \underline{17.99}
& 24.91 & \underline{22.66} & 18.11
& 21.64 & 32.5 & 57.6 & 34.10 & 35.68 & 32.54 & 21.27 & 20.27 \\

GaussianFormer-2$^{\dagger}$~\cite{huang2025gaussianformer2}
& 53.75 & 29.62
& 9.19 & \textbf{40.16} & \textbf{23.57}
& 39.62 & 45.71 & 16.97 & 25.13 & 20.54 & 17.83
& 21.65 & 32.88 & 58.27 & 34.18 & 40.29 & 34.77 & 22.69 & 20.05 \\

GPOcc++~\cite{zhou2026gpoccplus}
& \underline{54.11} & \underline{30.17}
& \underline{9.33} & 38.73 & 22.63
& 38.37 & \underline{46.79} & \textbf{18.98}
& \underline{25.75} & 20.79 & 17.87
& 21.96 & 30.27 & 60.56 & \textbf{37.05} & 42.37 & 36.12
& \underline{23.40} & \underline{21.93} \\

\midrule
\rowcolor{softteal}
\textbf{CGOcc (Ours)}
& \textbf{56.32} & \textbf{31.23}
& \textbf{10.53} & 38.21 & 21.80 & 40.30
& \textbf{48.82} & 17.66 & \textbf{26.25}
& 18.82 & 17.46 & 22.95 & \underline{34.09}
& \textbf{63.78} & 33.79 & \underline{43.91}
& \textbf{45.17} & \textbf{24.64} & \textbf{22.78} \\

\bottomrule
\end{tabular}%
}

\end{table*}

\begin{figure*}[!t]
    \centering
    \includegraphics[width=0.92\textwidth]{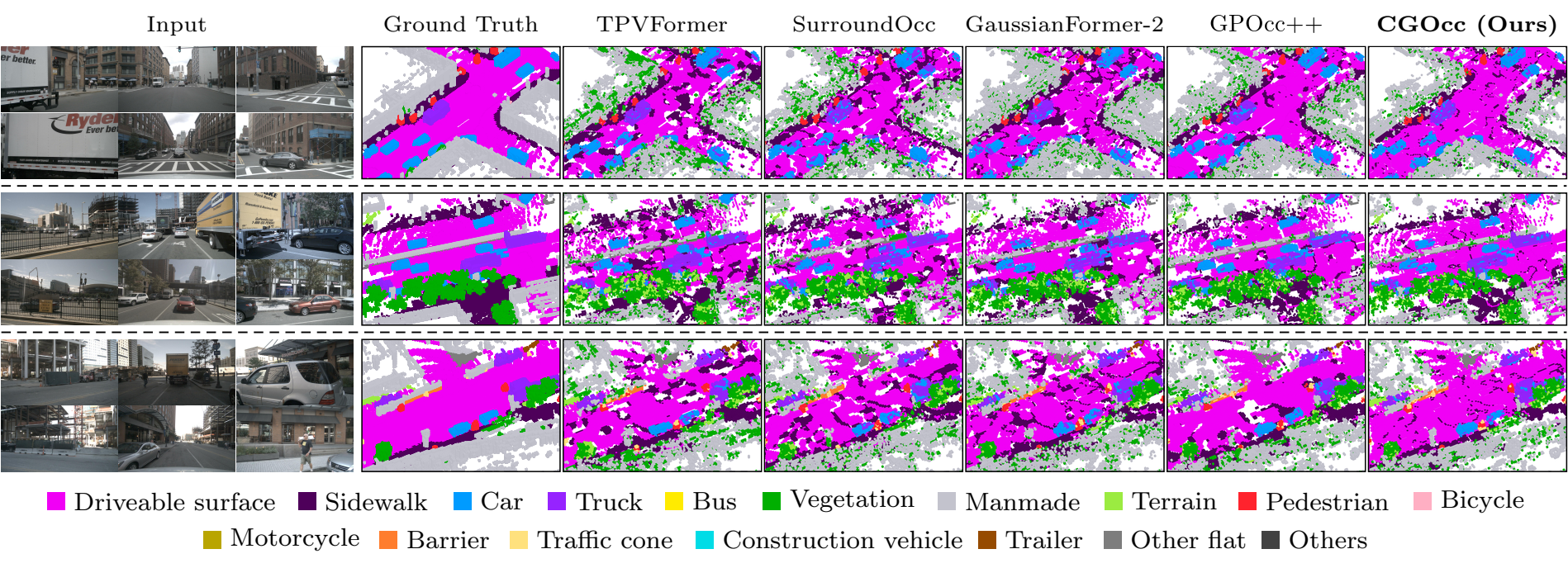}
    \caption{
        \textbf{Qualitative comparison on Occ3D-nuScenes validation set.}
    }
    \label{fig:qualitative_comparison}
\end{figure*}

\section{Experiments}

\subsection{Dataset and Metrics}
\label{sec:dataset_metrics}

We evaluate our method on Occ3D-nuScenes~\cite{tian2023occ3d},
which provides dense semantic occupancy annotations for
nuScenes~\cite{caesar2020nuscenes}.
Each input comprises RGB images from six surround-view cameras.
The prediction volume spans $[-40,40]$\,m along both horizontal axes
and $[-1,5.4]$\,m vertically, discretized into a
$200\times200\times16$ grid with a voxel size of $0.4$\,m.
The label space contains 17 occupied semantic classes and one free-space
class.
Evaluation is restricted to voxels marked as camera-visible by the
provided visibility mask~\cite{tian2023occ3d}.
We report mean intersection over union (mIoU) for semantic occupancy
and class-agnostic IoU for geometric occupancy, as commonly used in
semantic scene completion~\cite{cao2022monoscene}.
mIoU is averaged over the 17 occupied classes, whereas occupancy IoU
merges these classes into a single occupied category.
Both metrics are computed from the confusion matrix accumulated over
all evaluated frames, with zero IoU assigned to classes with zero union.
Additional evaluation under the SurroundOcc-nuScenes
protocol~\cite{wei2023surroundocc} is provided in the supplementary material.

\subsection{Implementation Details}
\label{sec:implementation_details}

We initialize 7,200 Gaussian candidates from Metric3Dv2~\cite{hu2024metric3dv2}
depth estimates and refine them in two decoder blocks utilizing features from
frozen DINOv2~\cite{oquab2024dinov2} and VGGT~\cite{wang2025vggt} encoders.
Our main model was trained for 30 epochs using AdamW~\cite{loshchilov2019adamw} on 8 NVIDIA RTX 5090 GPUs with a total batch size of 8.

We train CGOcc (Diag) independently under the same architecture and
parameter count as the full model. Its semantic correction is

\begin{equation}
  \mathbf r_v^{\mathrm{Diag}}=H_v(\mathbf m_v)-H_v(\mathbf 0).
  \label{eq:diag_readout_control}
\end{equation}

\noindent Where the diagonal descriptor $\mathbf m_v$ summarizes individual Gaussian contributions to each class.
\subsection{Main Results}
\label{sec:main_results}

\textbf{3D semantic occupancy results.}
As shown in Table~\ref{tab:gaussrender_occ3d}, CGOcc achieves the highest
OccIoU (56.32\%) and mIoU (31.23\%) on Occ3D-nuScenes, exceeding
GPOcc++~\cite{zhou2026gpoccplus} by 2.21 and 1.06
percentage points respectively. It leads in seven of 17 classes.
The largest gains are on terrain and truck (9.05 and 3.82
points, respectively), suggesting that depth initialization anchors
Gaussians and multi-view geometry refines their alignment with these
observed surfaces. CGOcc trails on bicycles and pedestrians by 0.83 and 1.97 points respectively, reflecting a larger proportion of predicted voxels from splat spillover around smaller objects.
In Fig.~\ref{fig:qualitative_comparison}, CGOcc leaves fewer road gaps and
stray vegetation labels on manmade structures (top row) and shows clearer
vehicle contours (middle row) than GaussianFormer-2~\cite{huang2025gaussianformer2} and
GPOcc++. These results demonstrate neighbor messages supply local
context for road continuity, and the class-conditional readout utilizes
dependencies between overlapping Gaussians to resolve boundary labels.
Sidewalk labels still spill onto the road (bottom row), where splats
can straddle the curb.

\begin{figure*}[!t]
    \centering
    \includegraphics[width=0.8\textwidth]{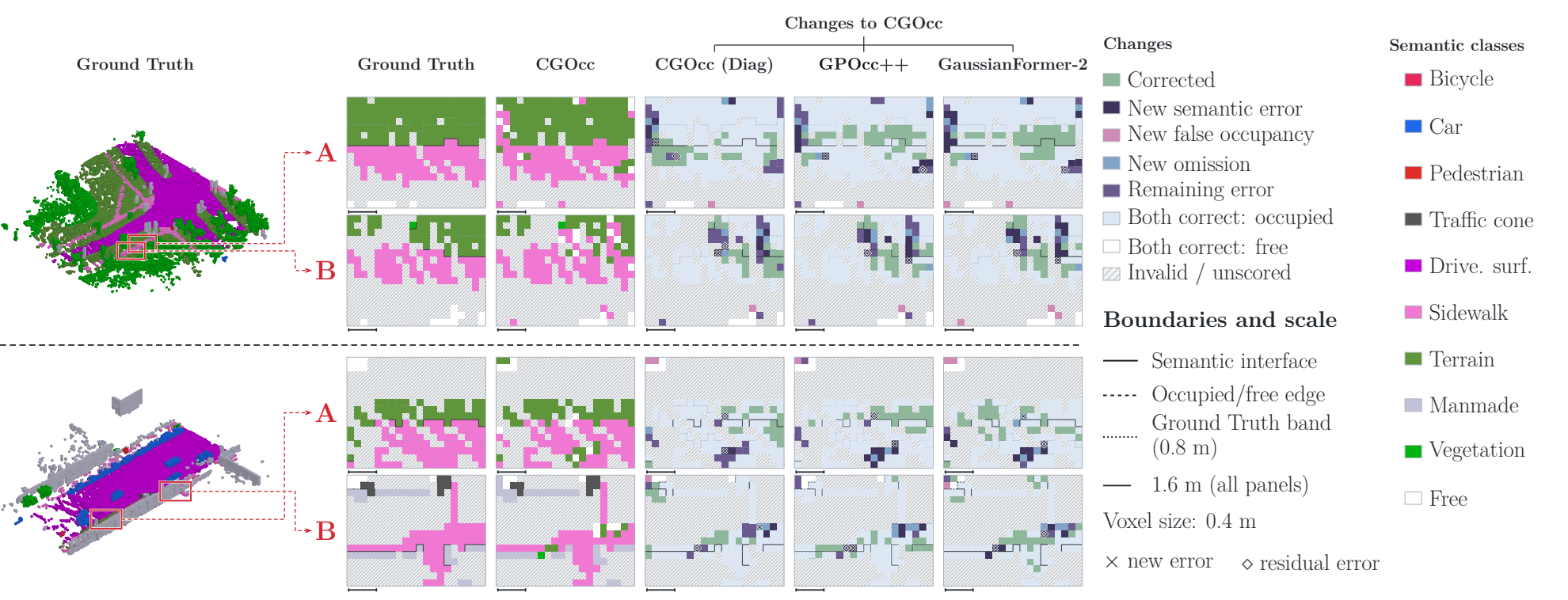}
    \caption{
        \textbf{Visualization of prediction changes near boundaries on
        Occ3D-nuScenes.}
        Local maps show changes from each comparison method to CGOcc.
        Corrections occur along semantic interfaces and occupied/free
        edges, alongside residual and new errors.
    }
    \label{fig:boundary_changes}
\end{figure*}

\begin{figure*}[!t]
    \centering
    \includegraphics[width=0.8\textwidth]{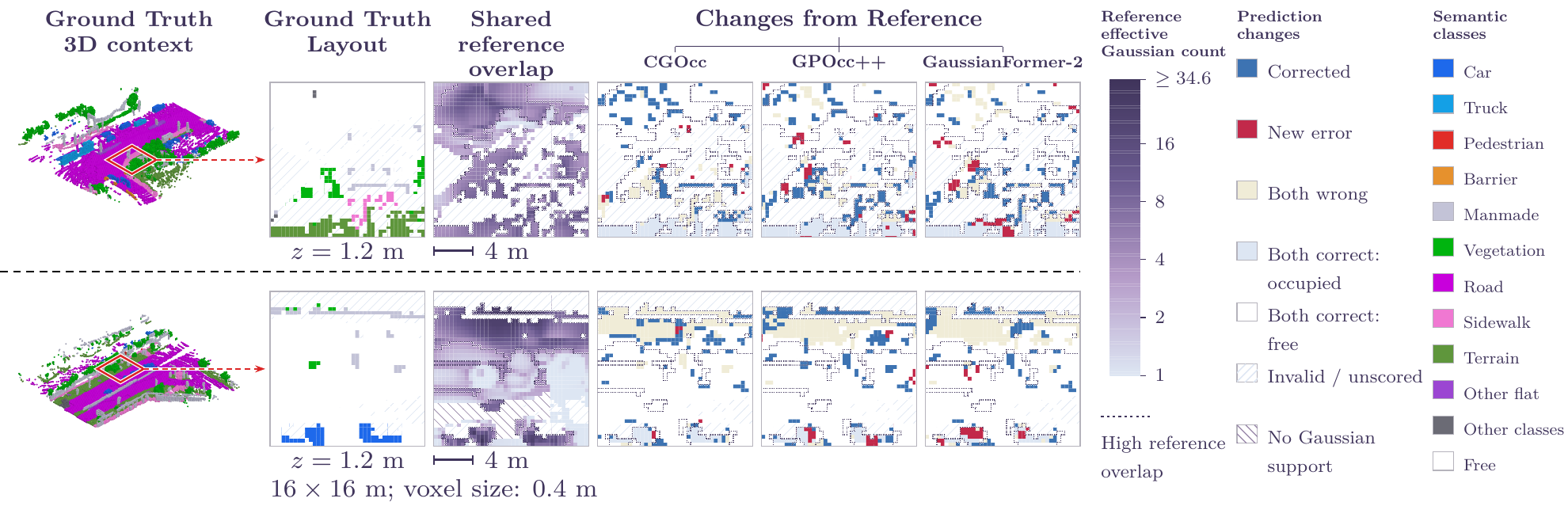}
    \caption{
        \textbf{Visualization of Gaussian overlap and prediction changes
        on Occ3D-nuScenes.}
        Overlap is measured using a shared fixed CGOcc (Diag) reference (elaborated in the supplementary material).
        Blue and red mark corrected predictions and new errors, respectively.
    }
    \label{fig:overlap_changes}
\end{figure*}

\begin{figure}[!t]
    \centering
    \includegraphics[width=0.95\linewidth]{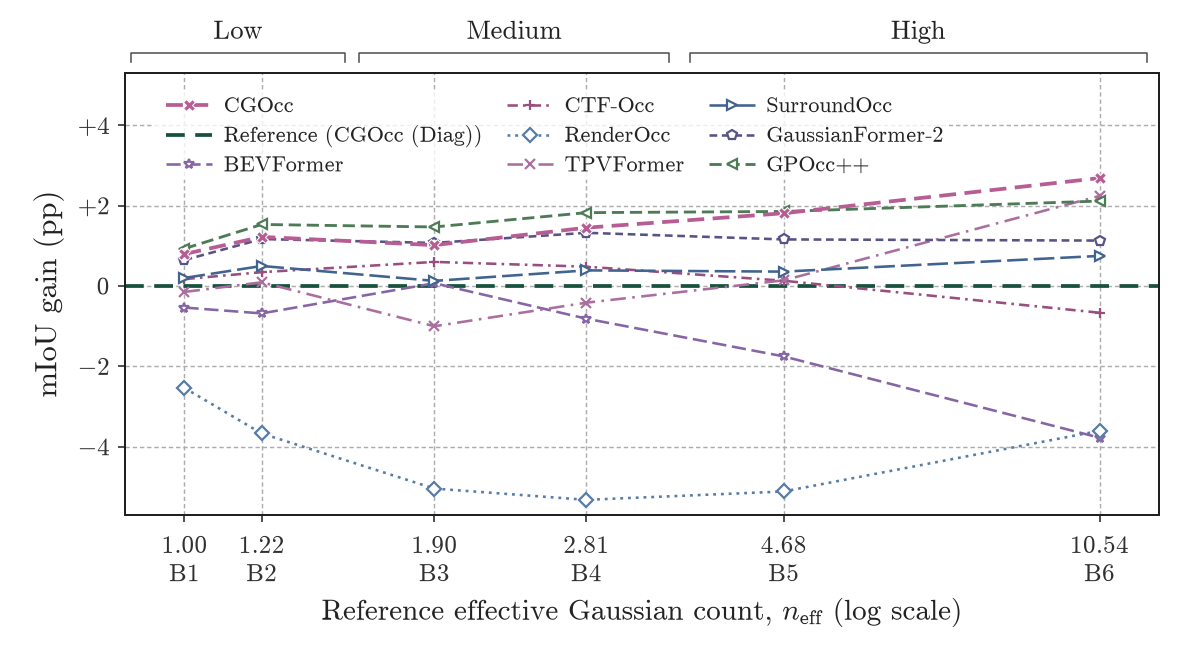}
    \caption{\textbf{Performance gains across reference-defined Gaussian overlap levels.} All curves use the same Gaussian overlap bins defined by CGOcc (Diag).
    }
    \label{fig:overlap_gain}
\end{figure}

\textbf{Semantic refinement near boundaries.}
In Fig.~\ref{fig:boundary_changes}, corrections trace the terrain--sidewalk
interface in crop A of the upper scene, suggesting that class-conditional
interactions help distinguish neighboring classes despite overlapping
Gaussian support. Corrections recur along this interface in all three comparisons.
In crop B of the lower scene, corrected voxels follow the
sidewalk outline near manmade structures, indicating that jointly
interpreting Gaussian evidence helps recover consistent labels across
neighboring locations. These corrections reduce confusion between the
sidewalk and nearby structures despite their overlapping Gaussian
representations. Residual errors in fragmented regions and isolated false
occupancy expose remaining difficulties with mixed class evidence and
Gaussian support extending into free space.
Distance-based comparisons, boundary definitions and distance binning are detailed
in the supplementary material.

\textbf{Semantic refinement across Gaussian overlap levels.}
In Fig.~\ref{fig:overlap_gain}, CGOcc gains mIoU over Diag across all
shared bins of the reference effective Gaussian count.
It trails GPOcc++ in B1--B4, nearly closes the gap in B5, and leads all
compared methods in B6, with its largest gain over Diag.
The advantage at high overlap supports the use of signed interactions,
weighted by both Gaussians' class probabilities, to resolve mixed
evidence from multiple primitives.
In Fig.~\ref{fig:overlap_changes}, corrections around the sidewalk and
vegetation in the upper crop show improved separation of adjacent
semantic regions. In the lower crop, CGOcc corrects shared error
locations along the elongated high-overlap band near a manmade
structure relative to both GPOcc++ and GaussianFormer-2.
These corrections manifest the benefit of reconciling overlapping
Gaussian contributions, while persistent and new errors reveal sensitivity
to ambiguous or incorrect class evidence.

\subsection{Ablation Study}

\textbf{Ablation of feature sources.}
To assess whether appearance and geometry provide complementary cues for occupancy prediction, we compare DINOv2~\cite{oquab2024dinov2},
VGGT~\cite{wang2025vggt}, and their combination
(Table~\ref{tab:ablation_feature_sources}).
VGGT alone outperforms DINOv2 alone on both metrics, while their
combination performs best.
The gain from adding appearance features suggests that they provide
semantic evidence complementary to geometric scene structure,
supporting their joint aggregation in the multi-source Gaussian
interaction module.

\begin{table}[!t]
    \centering
    \caption{\textbf{Ablation of feature sources.}}
    \label{tab:ablation_feature_sources}
    \small
    \setlength{\tabcolsep}{4pt}
    \renewcommand{\arraystretch}{1.12}
    \resizebox{\columnwidth}{!}{%
        \begin{tabular}{@{}lcccc@{}}
            \toprule
            Feature sources
            & DINOv2
            & VGGT
            & mIoU (\%) $\uparrow$
            & OccIoU (\%) $\uparrow$ \\
            \midrule
            DINOv2 only
            & $\checkmark$
            & $\times$
            & 28.71
            & 54.31 \\

            VGGT only
            & $\times$
            & $\checkmark$
            & 30.36
            & 55.22 \\

            DINOv2 + VGGT
            & $\checkmark$
            & $\checkmark$
            & \textbf{31.23}
            & \textbf{56.32} \\
            \bottomrule
        \end{tabular}%
    }
\end{table}

\begin{table}[!t]
    \centering
    \caption{\textbf{Ablation of depth initialization.}}
    \label{tab:ablation_depth_initialization}
    \small
    \setlength{\tabcolsep}{7pt}
    \renewcommand{\arraystretch}{1.15}
    \begin{tabular}{@{}lcc@{}}
        \toprule
        Initialization depth
        & mIoU (\%) $\uparrow$
        & OccIoU (\%) $\uparrow$ \\
        \midrule
        Uniform $z$-depth & 24.53 & 51.38 \\
        Metric3Dv2
        & \textbf{31.23}
        & \textbf{56.32} \\
        \bottomrule
    \end{tabular}
\end{table}

\textbf{Ablation of depth initialization.}
To test the decoder's reliance on near-surface Gaussian initialization,
we replace Metric3D v2~\cite{hu2024metric3dv2} depth with uniform
$z$-depth (Table~\ref{tab:ablation_depth_initialization}).
These drops reflect the role of scene-dependent depth in placing
Gaussians near observed surfaces and aligning multi-view feature
sampling with scene geometry.
Since center updates are bounded,
poor initial placement limits the decoder's ability to recover
surfaces and gather reliable semantic evidence.

\textbf{Decoder--covariance interaction.}
We examine how decoder mechanisms affect the mIoU advantage
of CGOcc over CGOcc (Diag) using eight paired training seeds
(Table~\ref{tab:diag_cov_synergy}).
W restricts observation attention to each backbone;
G removes geometry bias from query-attention logits;
N removes Gaussian-neighbor messages while retaining the FFN.
All inputs and other modules are retained.
CGOcc outperforms Diag in every configuration, with the
largest advantage observed under the full decoder.
These results support conditional synergy between decoder
mechanisms and covariance-aware readout, with neighbor messages
showing the largest mean interaction.
By combining observations and local context in Gaussian features,
the decoder may help the readout learn useful relationships
between primitives.

\begin{table}[t]
    \centering
    \caption{
        \textbf{Decoder--covariance interaction.}
        $S_j=\Delta_F-\Delta_j$, where
        $\Delta_j=Y_{j,\mathrm{Cov}}-Y_{j,\mathrm{Diag}}$
        and $Y$ denotes mIoU.
        Larger positive values indicate stronger synergy
        between decoder mechanism $j$ and covariance-aware prediction.
    }
    \label{tab:diag_cov_synergy}

    \begingroup
    \footnotesize
    \setlength{\tabcolsep}{3pt}
    \renewcommand{\arraystretch}{1.12}

    \resizebox{\columnwidth}{!}{%
        \begin{tabular}{@{}lccc@{}}
            \toprule
            Decoder variant & CGOcc (Diag) & CGOcc & $S_j$ \\
            \midrule
            Full (F)
            & $29.42 \pm 0.19$
            & $31.23 \pm 0.20$
            & --- \\

            Within-source (W)
            & $29.23 \pm 0.20$
            & $29.87 \pm 0.24$
            & $\bm{1.17 \pm 0.23}$ \\

            w/o geom.\ bias (G)
            & $29.11 \pm 0.17$
            & $30.08 \pm 0.24$
            & $\bm{0.84 \pm 0.19}$ \\

            w/o neighbor msg.\ (N)
            & $29.25 \pm 0.18$
            & $29.71 \pm 0.22$
            & $\bm{1.35 \pm 0.23}$ \\
            \bottomrule
        \end{tabular}%
    }
    \endgroup
\end{table}

\textbf{Ablation of semantic readout.}
Table~\ref{tab:readout_ablation} examines how overlapping Gaussians contribute
to voxel predictions. Diag, which summarizes individual contributions,
slightly improves in both metrics over the initial Gaussian-to-voxel prediction ($\mathbf P^{(0)}$-only).
Class-agnostic captures relationships between Gaussians using a normalized form of
{
\small
\begin{equation}
    \mathbf b_v
    =\left(\sum_g w_{vg}\boldsymbol\ell_g\right)
    \odot\left(\sum_g w_{vg}\boldsymbol\ell_g\right)
    -\sum_g w_{vg}^{2}
    (\boldsymbol\ell_g\odot\boldsymbol\ell_g).
    \label{eq:class_agnostic_interaction}
\end{equation}
}

Here $\odot$ denotes elementwise multiplication, and subtracting
self-contributions leaves interactions between distinct Gaussians.
Class-agnostic outperforms Diag, indicating that these relationships help
resolve voxel labels. Cov assigns each pair the weight
$w_{vg}w_{vh}p_{gc}p_{hc}$, so its contribution to class $c$ is small
when either Gaussian assigns that class low probability. Cov's higher
mIoU supports combining evidence from Gaussians that jointly support a class.

Cov (aligned factors) retains these weights but replaces
$\boldsymbol\ell_g^\top\boldsymbol\ell_h$ with
$\|\boldsymbol\ell_g\|_2\|\boldsymbol\ell_h\|_2$ in $X_{vc}$.
The result yields lower mIoU than full Cov, suggesting that relative
factor directions carry useful information by allowing contributions
to reinforce or offset one another, even at fixed class weights and
factor norms.

Finally, Cov (cross off) sets $\boldsymbol\xi_v$ to zero in the same
trained Cov checkpoint, recovering the learned reference $\mathbf B_v$.
The mIoU drop shows that cross statistics improve predictions by
reallocating probability among occupied classes while preserving
their total mass (Eq.~\eqref{eq:mass_preserving_prediction}).

\begin{table}[t]
    \centering
    \caption{
        \textbf{Ablation of semantic readout.}
        \textsuperscript{*} denotes Full model.
    }
    \label{tab:readout_ablation}

    \small
    \setlength{\tabcolsep}{8pt}
    \begin{tabular}{lcc}
        \toprule
        Readout Variant & mIoU (\%) $\uparrow$ & OccIoU (\%) $\uparrow$ \\
        \midrule
        $\mathbf P^{(0)}$-only & 29.25 & 54.56 \\
        Diag                  & 29.42 & 54.78 \\
        Class-agnostic        & 30.28 & 56.20 \\
        Cov (aligned factors) & 30.43 & 56.24 \\
        Cov\textsuperscript{*} & \textbf{31.23} & \textbf{56.32} \\
        \midrule
        Cov (cross off)       & 30.30 & 56.13 \\
        \bottomrule
    \end{tabular}
\end{table}

\begin{samepage}
\section{Conclusion}
\label{sec:conclusion}

We present CGOcc for semantic occupancy prediction, combining
geometry-conditioned multi-source Gaussian interaction with
class-conditional cross-Gaussian readout.
The readout refines occupied-class probabilities while preserving
the occupied mass and free-space probability of a learned reference.
With these designs, CGOcc achieves state-of-the-art performance on nuScenes while
spatial analyses show larger gains at high Gaussian overlap and improved predictions
near boundaries.

CGOcc remains susceptible to spillover at object boundaries. When
Gaussian splats extend beyond an object's surface into adjacent free
space, they can produce false occupancy and blur the predicted
contour. Future work could recalibrate occupied mass where depth
observations from multiple views agree on free space.

\end{samepage}

\section*{LLM Usage Statement}

We used OpenAI Codex to assist with language polishing,
grammar correction, and manuscript organization.
The tool also supported discussion and refinement of research
ideas, as well as identifying and correcting errors in the
implementation code.
The authors take full responsibility for the manuscript,
implementation, experimental results, and conclusions.

{
    \small
    \bibliographystyle{ieeenat_fullname}
    \bibliography{main}
}

\clearpage
\maketitlesupplementary

\appendix

\section{Additional Methodology Details}
\label{sec:supp_method_details}

\subsection{Initialization, Decoder, and Supervision}
\label{sec:supp_decoder_training}

We back-project fixed image references with bilinearly sampled Metric3D v2
depth. A valid initialization requires valid depth and an in-volume center.
Gaussians start with source-camera rotations, depth-dependent scales,
constant opacity, and zero read offsets. A learned projection initializes
queries from summed query and slot embeddings.

Each block samples at the center plus its read offset and again after
adding a $\tanh$-bounded query prediction scaled by $\lambda_{\mathrm{off}}$
times the clipped mean Gaussian scale. Sampling handles resizing, padding,
and projection validity. Source-level affine projections and pre-normalized
residual adapters align features before masked-softmax fusion of appearance
levels. VGGT levels remain separate. Camera, source, source-camera-membership,
and sampling-point embeddings identify observations.

Attention over these observations uses a scaled $\tanh$ geometry bias.
An encoder and head-specific mapping derive it from camera-axis depth,
distance, source-camera membership, validity, sampling identity, and
learned cues. For empty observation sets, queries proceed directly to the FFN.

Block 1 alone adds local graph attention. Any Gaussian can receive values
from initially valid non-self neighbors, whose attention weights use
learned relative-position bias. The graph jointly projects the weighted sum
and the unweighted mean of MLP-encoded relative positions normalized by $R$.
Message and FFN residuals share scale $\eta$ and require nonempty neighborhoods.

Shared heads refine Gaussian states. Center increments accumulate before
the bounded update, with $\mathbf a_g^{(0)}=\mathbf0$:
\begin{equation}
\begin{aligned}
\mathbf a_g^{(\ell+1)}&=\mathbf a_g^{(\ell)}+f_\mu(\mathbf q_g^{(\ell+1)}),\\
\boldsymbol\mu_g^{(\ell+1)}&=\boldsymbol\mu_g^{(0)}+B_g\tanh(\mathbf a_g^{(\ell+1)}).
\end{aligned}
\end{equation}
We apply bounded increments to read offsets, log scales, and opacity logits, with rotations updated by normalized quaternion composition of scaled axis-angle updates. Intermediate labels use floor-based voxel indexing at detached updated centers with valid initialization. Cross-entropy over in-volume, camera-visible occupied targets is normalized by their class-weight sum and empty target sets give zero loss. Terminal auxiliary targets require valid initialization and in-volume centers regardless of camera visibility. Only occupied targets enter the semantic loss.

\subsection{Dependency Encoding and Readout Heads}
\label{sec:supp_dependency_heads}

Gaussian inputs comprise geometry, opacity, class probabilities, entropy,
top-two margin, and valid observations pooled at final centers, with
fixed linear projection and normalization. We form the voxel context $\mathbf z_v$ from spatially pooled encodings,
class and factor means and their outer product, mean $d_g$, and semantic
variance. The context also includes $\log(1+S_v)$ for the total Gaussian
contribution to the voxel, $\log(1+n_v^{\mathrm{eff}})$ for the effective
number of contributing Gaussians, and $\log\mathbf P_v^{(0)}$ for the
initial voxel distribution over occupied classes and free space.
The reference and correction branches have separate encoders, each using
layer normalization and two 64-wide SiLU layers, followed by linear factor and
diagonal heads. nuScenes inputs and contexts have 45 and 263 dimensions.
Projection in Eq.~\eqref{eq:class_conditional_projection} costs
$O(ECr+VCr)$ for $E$ nonzero voxel--Gaussian pairs and $V$ voxels;
factor storage is $O(Nr+N)$. The positive-definite kernel permits signed
cross terms couple Gaussians with shared voxel support.

Independent MLPs $T_B,T_H$ use layer normalization and SiLU layers of
widths 128 and 64. For $T_v^B(\mathbf a,\mathbf b)
=T_B([\mathbf z_v^B;\mathbf a;\mathbf b])$, the reference adjustments are
\begin{equation}
\begin{aligned}
\mathbf t_v^B&=T_v^B(\mathbf m_v^B,\mathbf m_v^B)-T_v^B(\mathbf0,\mathbf0),\\
\mathbf s_v^B&=\beta_s\operatorname{ctr}[\tanh(W_s\mathbf t_v^B/\beta_s)],\\
b_v^B&=\beta_o\tanh(\mathbf w_o^\top\mathbf t_v^B/\beta_o).
\end{aligned}
\end{equation}
These enter Eq.~\eqref{eq:learned_reference_prediction}.
The correction head is
\begin{equation}
H_\psi(\mathbf z,\mathbf m,\mathbf t)
=QW_HT_H([\mathbf z;\mathbf m;\mathbf t]),
\end{equation}
where $Q$ is the fixed orthonormal Helmert basis for the zero-sum
subspace. Both MLPs receive 297-dimensional input vectors on nuScenes,
comprising a 263-dimensional voxel context and two
17-dimensional class-statistic vectors; bias-free projections
$W_s,\mathbf w_o,W_H$ are initialized to zero.

\subsection{Readout Controls and Properties}
\label{sec:supp_readout_controls}

Based on Eq.~\eqref{eq:diag_readout_control}, CGOcc (Diag) retains both encoders, all factor/diagonal heads, and the full
correction head.
In Diag, $D_{vc}$ retains the self-contributions from $LL^\top$.
Its architecture and parameter count match Cov, with all task-specific
modules jointly trained. Cross off instead fixes a Cov checkpoint and
zeros only $\boldsymbol\xi_v$, recovering its own reference $\mathbf B_v$.

Class-agnostic normalizes the vector in
Eq.~\eqref{eq:class_agnostic_interaction} using
\begin{equation}
\begin{aligned}
D_v^{\mathrm{agn}}&=\sum_gw_{vg}^2(d_g+\|\boldsymbol\ell_g\|_2^2),\\
\eta_{vj}&=\frac{b_{vj}}{D_v^{\mathrm{agn}}+\epsilon_X+|b_{vj}|}.
\end{aligned}
\end{equation}
The full $r$-vector replaces $\boldsymbol\xi_v$ in the shared-head
difference, changing the input dimension from $d_z+2C$ to $d_z+C+r$.
The variant retains the hidden layers, reference, context, and
class-conditional diagonal descriptors. We train it independently
with the same objective, masks, and schedule. The comparison evaluates
the combined change in class weighting and the organization of
cross-Gaussian statistics.

Both corrections follow Eqs.~\eqref{eq:conditional_cross_correction}
and~\eqref{eq:mass_preserving_prediction}. The centered $\tanh$
gives zero-sum corrections bounded by $\tau$. Normalized reweighting
preserves occupied mass and free probability, potentially changing
occupied/free labels under the $(C+1)$-way argmax. Zero incremental
input cancels the shared-head difference, recovering $\mathbf B_v$.
Unsupported, single-contributor, and zero-mass voxels use this
reference, which equals $\mathbf P_v^{(0)}$ with at most one contributor.

\subsection{Implementation Settings}
\label{sec:supp_implementation_settings}

We precompute the outputs of DINOv2, VGGT, and Metric3D v2. Images are
resized to $504\times896$ for DINOv2 and padded and resized to
$518\times518$ for VGGT. We initialize the query embeddings and appearance
frontend from the released GaussTR checkpoint~\cite{jiang2025gausstr} and
keep the appearance neck frozen.

Table~\ref{tab:supp_hyperparameters} lists the architecture and readout
settings on nuScenes. Learned scalar initializations are marked ``init.'';
other entries are fixed. In meters, valid depth is $(0.1,51.2]$ and
\begin{equation}
\begin{aligned}
s_{g,\mathrm{tan}}^{(0)}&=\operatorname{clip}(0.01z_g^{(0)},0.10,0.40),\\
B_g&=\operatorname{clip}(0.30z_g^{(0)},0.30,5.0).
\end{aligned}
\end{equation}
Both tangential axes use this scale, with axial scale and opacity initially
$0.10$\,m and $0.5$. Refinement bounds read-offset, log-scale, and
opacity-logit increments by $B_g$, $1.5$, and $2.0$, with axis-angle
updates scaled by $0.35$. We clip projection scales and sampling mean
scales to $[0.05,2.5]$\,m. Cubic splat stencils use three times the largest
world-axis standard deviation, rounded up in voxel units and capped at
16 voxels before clipping to the volume. Readout heads require
$S_v>10^{-12}$ and $n_v^{\mathrm{eff}}>1+10^{-6}$.

\begin{table}[H]
\centering
\caption{\textbf{Architecture and readout settings on nuScenes.}}
\label{tab:supp_hyperparameters}

\small
\setlength{\tabcolsep}{2pt}
\begin{tabular}{@{}lr@{}}
\toprule
Setting & Value \\
\midrule
Gaussians / blocks / attention heads & 7,200 / 2 / 4 \\
Hidden / attention FFN width & 256 / 512 \\
Adapter / geom.\ bias hidden width & 256 / 64 \\
Dropout / sampling points / slots & 0 / 2 / 48 \\
Sampling-offset scale $\lambda_{\mathrm{off}}$ & 0.5 \\
Graph neighbors $K$ / radius $R$ & 16 / 1.6\,m \\
Factor rank $r$ / correction bound $\tau$ & 8 / 0.5 \\
$d_{\min}$ / $\epsilon_X$ / $\alpha$ & 0.02 / 0.02 / $10^{-4}$ \\
Reference scales $\beta_s$ / $\beta_o$ & 0.25 / 0.25 \\
Gating $\gamma$ / geometry bias $\beta$ (init.) & 1.0 / 5.0 \\
Graph $\eta$ / adapter residual scale (init.) & 0.1 / 0.05 \\
\bottomrule
\end{tabular}
\end{table}

We train the main model on eight NVIDIA RTX 5090 GPUs with an AdamW
weight decay of $0.01$. The peak learning rate is $10^{-3}$ by default,
$3\times10^{-4}$ for geometry updates, and $2\times10^{-4}$ for the
dependency encoder and prediction heads. The 30-epoch cosine schedule
begins with 1.5 epochs of linear warm-up. Neural blocks run in FP16,
while geometry, sparse aggregation, and loss computation remain in FP32.

\section{Boundary and Overlap Evaluation Protocols}
\label{sec:supp_evaluation}

\subsection{Evaluation Setup and Metrics}
\label{sec:supp_eval_domain}

The regional analyses use the validation set from
Occ3D-nuScenes dataset~\cite{tian2023occ3d}. Let $y(v)$ denote the
label of voxel $v$, $\mathcal K=\{0,\ldots,16\}$ the occupied
classes, and $f=17$ free space. Given the label set $\mathcal Y=\mathcal K\cup\{f\}$ and the
camera-visibility mask $M$, the scoring domain is
$\mathcal V=\{v:M(v)=1,\ y(v)\in\mathcal Y\}$.
All methods in a regional comparison share the same target frames,
labels, scoring mask, and region assignments.

For model $m$ and region $\mathcal R_t\subseteq\mathcal V_t$ in
frame $t$, we first accumulate an $18\times18$ confusion matrix:
\begin{equation}
H_{ij}^{(m,\mathcal R)}
=\sum_t\sum_{v\in\mathcal R_t}a_{t,ij}^{(m)}(v),
\label{eq:supp_region_confusion}
\end{equation}
where $a_{t,ij}^{(m)}(v)=1$ when $y_t(v)=i$ and
$\hat y_{m,t}(v)=j$, and zero otherwise. Rows denote ground truth
and columns predictions. Omitting the superscripts, the class union
$U_k$ and semantic mIoU are computed as
\begin{equation}
\begin{aligned}
U_k&=\sum_jH_{kj}+\sum_iH_{ik}-H_{kk},\\
\operatorname{mIoU}(H)
&=\frac{100}{17}\sum_{k=0}^{16}\frac{H_{kk}}{U_k},
\end{aligned}
\label{eq:supp_region_miou}
\end{equation}
where a zero-union class contributes zero. Free space is excluded
from the class average but retained in the confusion matrix, so
false occupied predictions on free voxels contribute to semantic
false positives. OccIoU merges all occupied classes:
\begin{equation}
    \operatorname{OccIoU}(H)=\frac{100\,TP}{TP+FP+FN}.
\end{equation}
In the boundary-distance evaluation below, OccIoU is undefined for
bins with $TP+FP+FN=0$.

\subsection{Boundary Definition and Distance Binning}
\label{sec:supp_boundary_protocol}

Let $L$ be the LiDAR-validity mask and $\mathcal N_6(v)$ the six
face-sharing neighbors of $v$. The seed set $\mathcal S$ contains both
endpoints of every pair with $u\in\mathcal N_6(v)$,
$L(v)=L(u)=1$, and $y(v)\ne y(u)$.
Define Boundary as $\mathcal B=\mathcal V\cap\mathcal S$ and
Interior as $\mathcal I=\mathcal V\setminus\mathcal S$.
The seed set $\mathcal S$ includes interfaces between different
occupied classes and between occupied and free voxels. LiDAR validity determines whether a label
transition is confirmed, while camera visibility determines scoring.
Boundary $\mathcal B$ and Interior $\mathcal I$ partition $\mathcal V$,
with Interior comprising the occupied and free voxels outside the
confirmed boundary, including those adjacent to unknown regions.

For voxel size $h=0.4$\,m, the unsigned 3D distance is
\begin{equation}
d(v)=h\min_{u\in\mathcal S}\|v-u\|_2,
\label{eq:supp_boundary_distance}
\end{equation}
with display neighborhood
$\mathcal N_{0.8}=\{v\in\mathcal V:d(v)\leq0.8\,\mathrm m\}$.
If $\mathcal S$ is empty, $d(v)=+\infty$. Distances are computed
on the full volume before slicing or cropping. The 0.8\,m
neighborhood provides display context and contains some Interior voxels. Distance curves
use the bin index
\begin{equation}
    \nu(v)=\left\lfloor d(v)/h+1/2\right\rfloor,
\end{equation}
with plotted centers
$0,0.4,\ldots,4.0$\,m. The zero-distance bin
contains exactly the scored boundary seeds, and a separate tail bin
retains distances of at least $4.2$\,m.

\subsection{Gaussian Overlap Measurement and Binning}
\label{sec:supp_overlap_protocol}

Overlap is measured using a shared fixed CGOcc (Diag) reference. For its spatial
contributions $\omega_{vg}$, define
\begin{equation}
\begin{gathered}
S_v=\sum_g\omega_{vg},\qquad w_{vg}=\omega_{vg}/S_v,\\
n_v^{\mathrm{eff}}=
\left(\sum_gw_{vg}^2\right)^{-1}.
\end{gathered}
\label{eq:supp_effective_overlap}
\end{equation}
We form overlap groups on
$\mathcal V^+=\{v\in\mathcal V:S_v>10^{-12}\}$, with supported free voxels
included and unsupported voxels recorded separately.

We pool $n_v^{\mathrm{eff}}$ over all supported valid voxels
in the entire training set and estimate thresholds using linearly
interpolated empirical quantiles:
\begin{equation}
    q_j=Q_{j/6}(n^{\mathrm{eff}}),\qquad j=1,\ldots,5.
\end{equation}
With $q_0=-\infty$ and $q_6=+\infty$, group $\mathcal O_j$ contains
voxels $v\in\mathcal V^+$ satisfying
$q_{j-1}<n_v^{\mathrm{eff}}\leq q_j$.
Voxels at a threshold fall in the lower group. All models share these
assignments, and we plot each group at its pooled evaluation-set median
$n_v^{\mathrm{eff}}$. Pairing adjacent groups gives Low $(1,2)$,
Medium $(3,4)$, and High $(5,6)$.

\subsection{Visualization Settings and Error Definitions}
\label{sec:supp_visualization}

Corresponding panels share slice coordinates, crop windows, voxel
size, and visibility masks. Each local cell represents one voxel and
invalid locations are masked, with boundary interfaces and distance fields are computed before cropping.
Solid and dashed
lines mark occupied-class and occupied/free interfaces, respectively;
dotted lines mark the 0.8\,m neighborhood. Local change maps cover
all scored voxels in the crop, including its non-boundary portion.

For comparison labels $b(v)$ and Cov labels $c(v)$, Corrected means
$b(v)\ne y(v)$ and $c(v)=y(v)$. The two-color overlap maps mark
New error only when $b(v)=y(v)$ and $c(v)\ne y(v)$.
The detailed boundary maps distinguish new semantic errors,
which change a correct occupied class to another occupied class,
and new false occupancy, which changes correct free space to
occupied. A new omission requires
$y(v)\in\mathcal K$, $b(v)\in\mathcal K$, and $c(v)=f$.
Remaining error contains Cov errors outside these three categories, while
both-correct voxels are separated into occupied and free.

Overlap heatmaps use the common logarithmic mapping
\begin{equation}
t(v)=\operatorname{clip}_{[0,1]}
\left(\frac{\log n_v^{\mathrm{eff}}}{\log U}\right),
\label{eq:supp_overlap_color}
\end{equation}
with $U\approx34.58565$, the fixed training-calibration 99th
percentile. Darker purple denotes greater overlap and dotted purple
contours enclose the High group. Quantitative gain curves connect
measured group values by straight segments and
correction counts describe changes in label accuracy.

\section{Extended Experimental Results}
\label{sec:supp_extended_results}

\begin{figure}[H]
    \centering
    \includegraphics[width=\columnwidth]{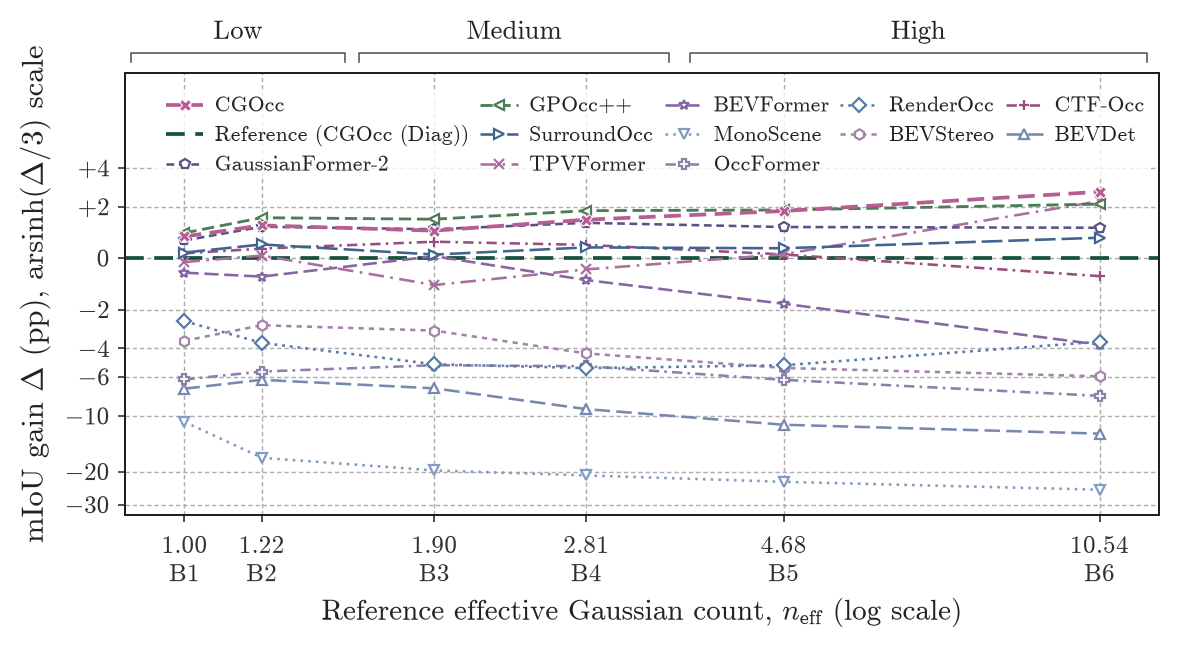}
    \caption{
    \textbf{Full comparison across reference-defined Gaussian overlap
    levels on Occ3D-nuScenes.}
    All methods use the same voxel bins defined by CGOcc (Diag).
    The ordinate plots $\operatorname{arsinh}(\Delta/3)$, with tick
    labels giving the signed mIoU difference $\Delta$ relative to
    CGOcc (Diag) in percentage points. The logarithmic horizontal
    axis uses the median reference effective Gaussian count in each bin.
    }
    \label{fig:full_overlap_gain}
\end{figure}

\begin{figure*}[!t]
    \centering

        \includegraphics[
            width=\textwidth,
            height=0.84\textheight,
            keepaspectratio
        ]{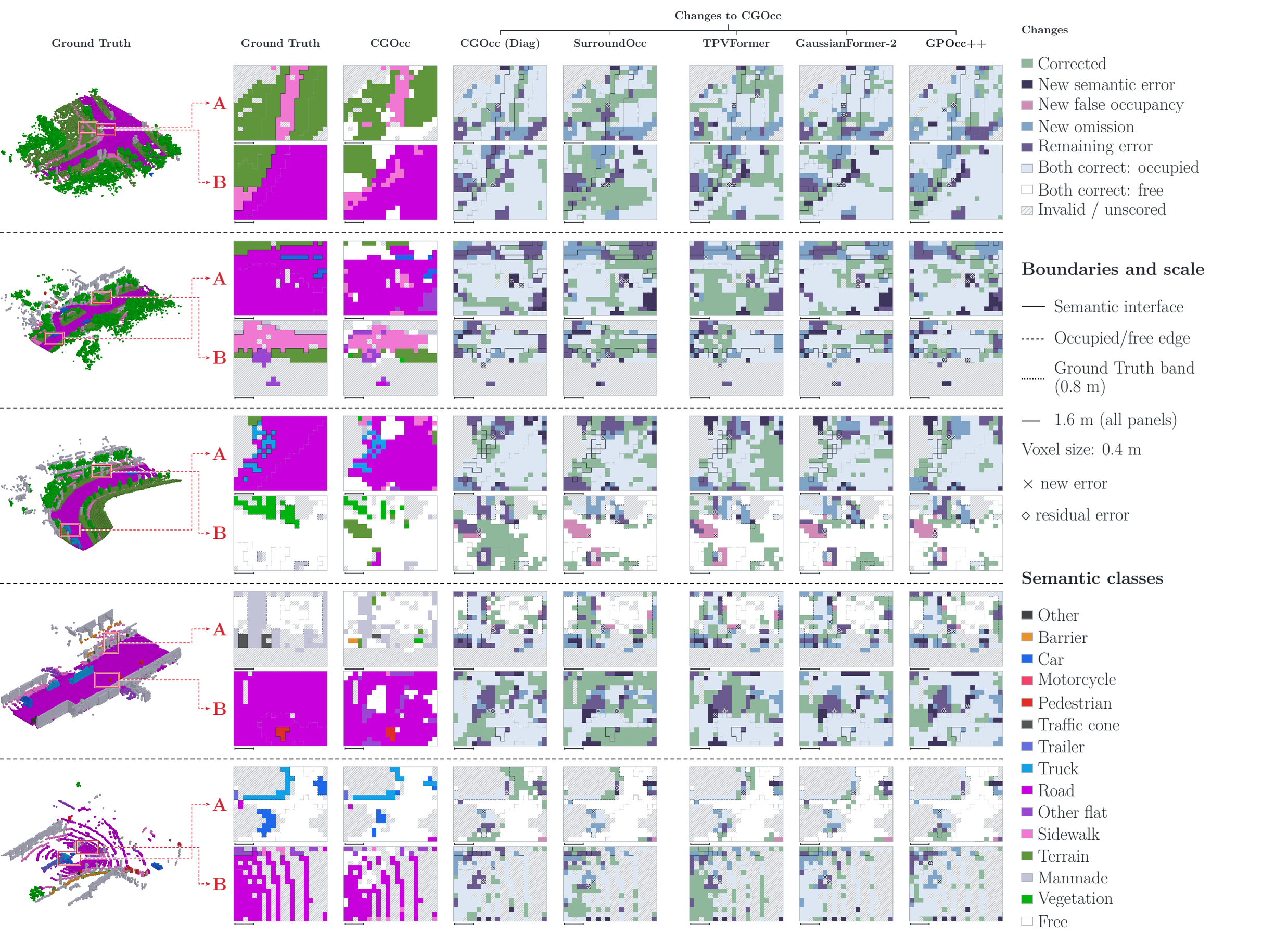}

    \caption{
    \textbf{\textbf{Visualization of prediction changes near boundaries
    in five additional Occ3D-nuScenes scenes.}}
    Each scene shows two local regions, with change maps comparing
    CGOcc against CGOcc (Diag) and external baselines.
    Solid and dashed lines mark occupied-class and occupied/free
    interfaces, respectively; dotted lines indicate the 0.8\,m
    boundary neighborhood.
    Colors distinguish corrected voxels and different error types.
    }
    \label{fig:full_boundary_changes}
\end{figure*}

\begin{figure*}[!t]
    \centering
    \includegraphics[
        width=\textwidth,
        height=0.84\textheight,
        keepaspectratio
    ]{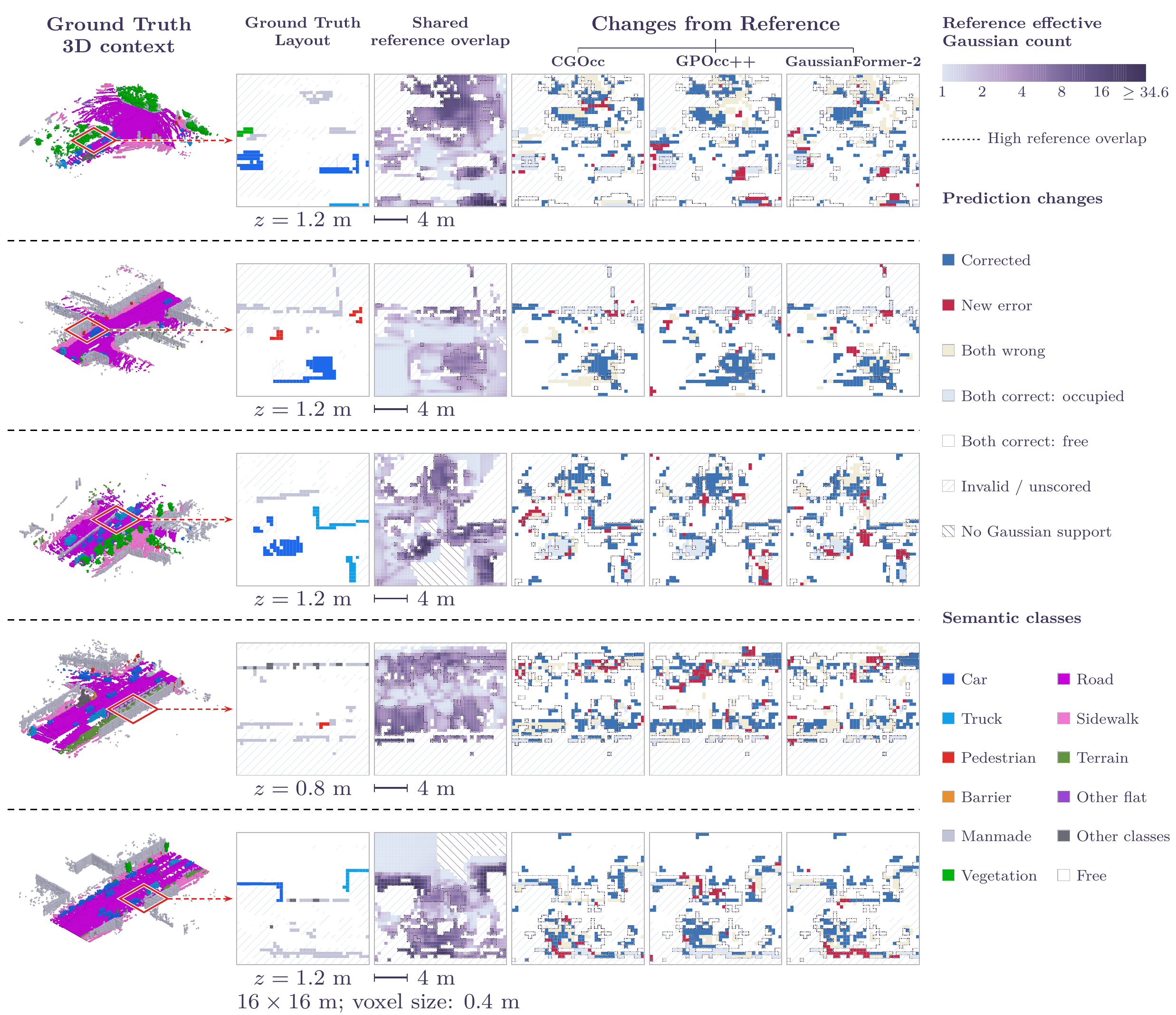}

    \caption{
    \textbf{Visualization of Gaussian overlap and prediction changes
    in five additional Occ3D-nuScenes scenes.}
    The change maps compare predictions from CGOcc, GPOcc++, and GaussianFormer-2 relative to a fixed CGOcc (Diag) reference. The overlap maps show the effective Gaussian contribution count from that reference.
    Darker purple indicates greater overlap, and dotted contours
    mark high-overlap regions.
    In the change maps, blue and red denote corrected predictions
    and new errors, respectively.
    }
    \label{fig:full_overlap_changes}
\end{figure*}

We report quantitative and qualitative results under the
SurroundOcc-nuScenes evaluation protocol
(Table~\ref{tab:surroundocc_nuscenes} and
Fig.~\ref{fig:surroundocc_results}), followed by extended boundary and
Gaussian-overlap analyses under the Occ3D-nuScenes evaluation
protocol (Figs.~\ref{fig:full_boundary_changes},
\ref{fig:full_overlap_gain}, and~\ref{fig:full_overlap_changes}).

\textbf{SurroundOcc evaluation protocol.} For Table~\ref{tab:surroundocc_nuscenes} and
Fig.~\ref{fig:surroundocc_results}, we use the dense semantic
occupancy annotations released by
SurroundOcc~\cite{wei2023surroundocc} for nuScenes.
Each input comprises RGB images from six surround-view cameras.
The prediction volume spans $[-50,50]$\,m along both horizontal
axes and $[-5,3]$\,m vertically, discretized into a
$200\times200\times16$ grid with a voxel size of $0.5$\,m.
The label space contains 16 occupied semantic classes and one
free-space class.
Following the official evaluation protocol, voxels assigned the
ignore label are excluded, and no additional camera-visibility
mask is applied.
We report mIoU averaged over the 16 occupied classes and OccIoU
obtained by merging these classes into a single occupied category.
Both metrics are computed from intersection and union counts
accumulated over all evaluated frames.

\textbf{Quantitative and qualitative results on SurroundOcc-nuScenes.}
As shown in Table~\ref{tab:surroundocc_nuscenes}, CGOcc leads with
34.65\% OccIoU and 21.95\% mIoU, exceeding GPOcc++~\cite{zhou2026gpoccplus}
by 1.17 points in mIoU. Its largest per-class gains are on buses and cars.
In Fig.~\ref{fig:surroundocc_results}, the bus retains its elongated
footprint (fourth row), consistent with depth initialization and multi-view
refinement keeping Gaussians close to the vehicle surface. At its outline,
overlapping splats can contribute to both bus and road labels. The
class-conditional readout can reweight these labels using evidence from
pairs of Gaussians that jointly support a class. At the intersection
(second row), CGOcc leaves fewer road gaps than
OccFormer~\cite{zhang2023occformer} and
GaussianFormer-2~\cite{huang2025gaussianformer2}, suggesting that neighbor
messages help Gaussians with weak local observations draw on evidence
from nearby parts of the road.
The final row shows a largely continuous road beside a fragmented
sidewalk, underscoring the difficulty of preserving narrow structures
when a splat covers voxels on both sides of the curb. A few misplaced
voxels can likewise lower IoU for pedestrians and traffic cones, on
which CGOcc trails SurroundOcc~\cite{wei2023surroundocc}.

\begin{figure*}[!t]
    \centering
    \begin{minipage}[t]{0.49\textwidth}
        \centering
        \includegraphics[width=\linewidth]
        {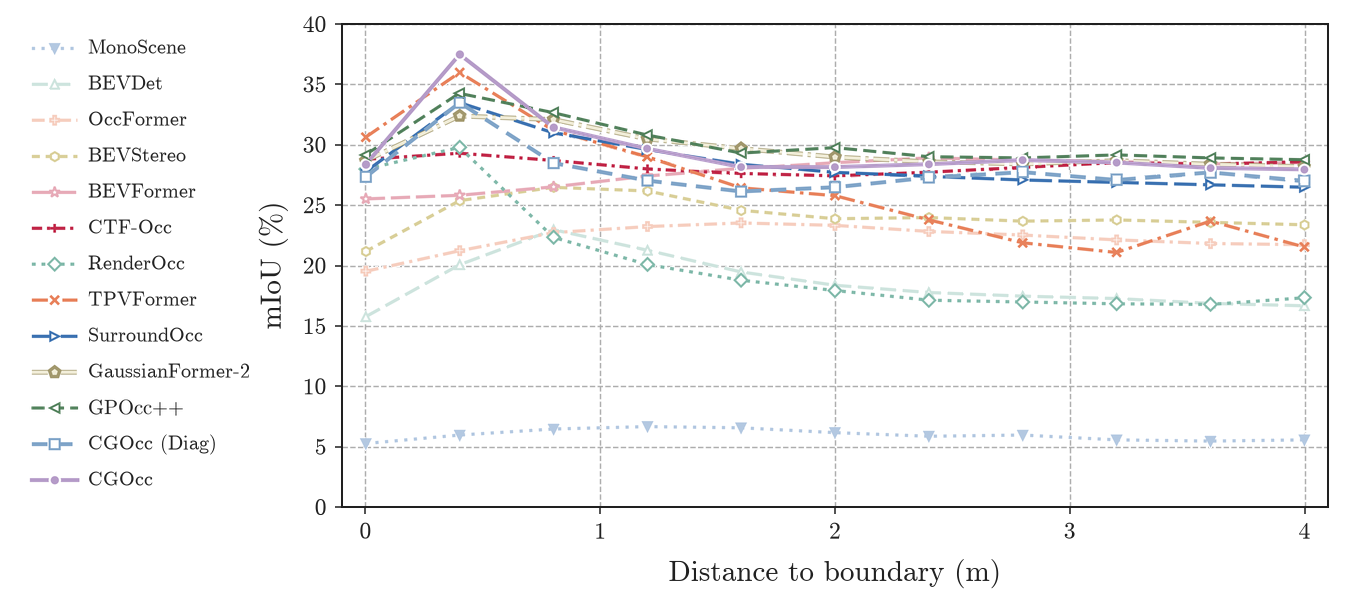}
        \par\smallskip
        {\small (a) mIoU\par}
    \end{minipage}\hfill
    \begin{minipage}[t]{0.49\textwidth}
        \centering
        \includegraphics[width=\linewidth]
        {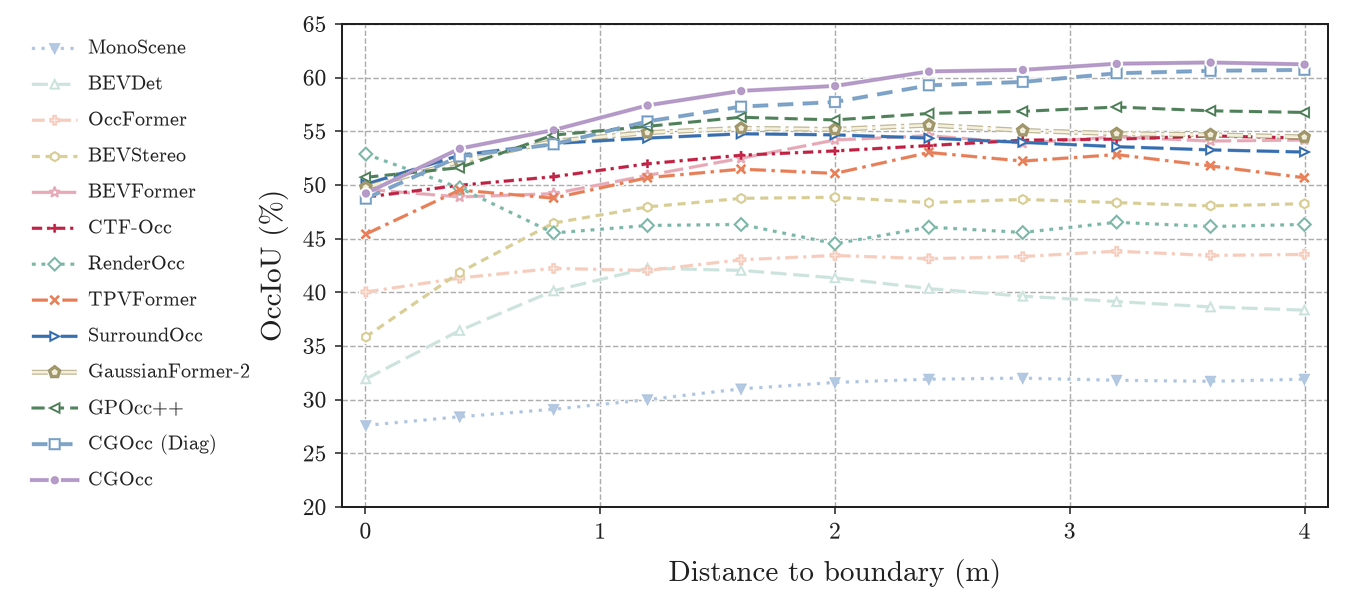}
        \par\smallskip
        {\small (b) OccIoU\par}
    \end{minipage}

    \caption{
    \textbf{Quantitative comparison across distances to
    ground-truth boundaries.}
    Panels (a) and (b) report mIoU and OccIoU over common distance bins.
    For each scored voxel, the unsigned 3D Euclidean distances are measured from its center to the center of the nearest ground-truth boundary seed voxel.
    }
    \label{fig:boundary_distance}
\end{figure*}

\textbf{Extended boundary analysis.}
\phantomsection
\label{par:supp_extended_boundary}
As shown in Fig.~\ref{fig:boundary_distance}a, CGOcc achieves the highest
mIoU at 0.4\,m from the boundary, exceeding CGOcc (Diag) by about four
points. The gap generally narrows farther from the boundary. Both models
use the same decoder architecture, but CGOcc also considers which pairs
of Gaussians support the same class. The larger gain near boundaries
suggests that this helps most where Gaussians from neighboring surfaces
predict different labels for the same voxel. CGOcc also leads in OccIoU
at every nonzero distance shown, although RenderOcc scores higher at the
boundary itself (Fig.~\ref{fig:boundary_distance}b).

In the first scene of Fig.~\ref{fig:full_boundary_changes}, CGOcc corrects
errors made by Diag along the narrow sidewalk where it borders terrain
(crop~A) and road (crop~B). At the curb, a Gaussian covering the road can
extend onto the sidewalk. When two overlapping Gaussians both predict
sidewalk, CGOcc can give that label more weight and preserve the narrow
strip. In the final scene, CGOcc recovers parts of the vehicle outlines
in crop~A and narrow road segments in crop~B. Some car voxels along the
outline are still labeled as free. These gaps show that recovering the
full extent of a vehicle remains difficult even with depth-guided
Gaussian placement.

\textbf{Extended Gaussian-overlap analysis.}
As shown in Fig.~\ref{fig:full_overlap_gain}, CGOcc outperforms
CGOcc (Diag) in every overlap bin. CGOcc trails GPOcc++ at low and
medium overlap, nearly closes the gap in B5, and ranks first in B6,
where its gain over Diag reaches about 2.7 mIoU points. When one
Gaussian dominates a voxel, the others carry little weight in the
class prediction. At high overlap, several Gaussians can predict
different classes for the same voxel. The larger gain over Diag
suggests that CGOcc resolves some of these disagreements by giving
more weight to labels supported by pairs of Gaussians.

Figure~\ref{fig:full_overlap_changes} compares CGOcc, GPOcc++, and
GaussianFormer-2 against the same Diag reference. In the second scene,
all three methods recover a cluster of car voxels that Diag mislabels
in the lower right of the crop. For CGOcc, pairs of overlapping
Gaussians that both predict car can reinforce that label across
neighboring voxels on the vehicle. In the fourth scene, CGOcc also
corrects labels along the lower band of manmade structures. The upper
band has high overlap as well, yet new errors appear beside the
corrections. This suggests that even where several Gaussians cover
a voxel, they can still favor the wrong class.

\begin{table*}[!t]
\centering
\caption{\textbf{3D semantic occupancy prediction results on the
SurroundOcc-nuScenes validation set.}
$^{\dagger}$GaussianFormer-2 uses the 128-channel configuration
($\mathrm{Ch.}=128$). Column-wise best results are \textbf{bold} and second best results are \underline{underlined.}}
\label{tab:surroundocc_nuscenes}

\begingroup
\small
\setlength{\tabcolsep}{3pt}
\renewcommand{\arraystretch}{1.08}
\resizebox{\textwidth}{!}{%
\begin{tabular}{l|cc|*{16}{c}}
\toprule
Method & OccIoU $\uparrow$ & mIoU $\uparrow$
& \occclass{FF7832}{barrier}
& \occclass{FFB4C8}{bicycle}
& \occclass{FFFF00}{bus}
& \occclass{0096FF}{car}
& \occclass{00FFFF}{const. veh.}
& \occclass{B4B400}{motorcycle}
& \occclass{FF0000}{pedestrian}
& \occclass{FFF096}{traffic cone}
& \occclass{964600}{trailer}
& \occclass{A032FF}{truck}
& \occclass{FF00FF}{drive. surf.}
& \occclass{8C8C8C}{other flat}
& \occclass{4B004B}{sidewalk}
& \occclass{96F03C}{terrain}
& \occclass{E6E6FF}{manmade}
& \occclass{00AF00}{vegetation} \\
\midrule

MonoScene~\cite{cao2022monoscene}
& 23.96 & 7.31
& 4.03 & 0.35 & 8.00 & 8.04 & 2.90 & 0.28 & 1.16 & 0.67
& 4.01 & 4.35 & 27.72 & 5.20 & 15.13 & 11.29 & 9.03 & 14.86 \\

Atlas~\cite{murez2020atlas}
& 28.66 & 15.00
& 10.64 & 5.68 & 19.66 & 24.94 & 8.90 & 8.84 & 6.47 & 3.28
& 10.42 & 16.21 & 34.86 & 15.46 & 21.89 & 20.95 & 11.21 & 20.54 \\

BEVFormer~\cite{li2022bevformer}
& 30.50 & 16.75
& 14.22 & 6.58 & 23.46 & 28.28 & 8.66 & 10.77 & 6.64 & 4.05
& 11.20 & 17.78 & 37.28 & 18.00 & 22.88 & 22.17 & 13.80 & 22.21 \\

TPVFormer-lidar~\cite{huang2023tpvformer}
& 11.51 & 11.66
& 16.14 & 7.17 & 22.63 & 17.13 & 8.83 & 11.39 & 10.46 & 8.23
& 9.43 & 17.02 & 8.07 & 13.64 & 13.85 & 10.34 & 4.90 & 7.37 \\

TPVFormer~\cite{huang2023tpvformer}
& 30.86 & 17.10
& 15.96 & 5.31 & 23.86 & 27.32 & 9.79 & 8.74 & 7.09 & 5.20
& 10.97 & 19.22 & 38.87 & 21.25 & 24.26 & 23.15 & 11.73 & 20.81 \\

OccFormer~\cite{zhang2023occformer}
& 31.39 & 19.03
& 18.65 & 10.41 & 23.92 & 30.29 & 10.31 & 14.19
& \underline{13.59} & 10.13
& 12.49 & 20.77 & 38.78 & 19.79 & 24.19 & 22.21 & 13.48 & 21.35 \\

GaussianFormer~\cite{huang2024gaussianformer}
& 29.83 & 19.10
& 19.52 & 11.26 & 26.11 & 29.78 & 10.47 & 13.83 & 12.58 & 8.67
& 12.74 & 21.57 & 39.63 & 23.28 & 24.46 & 22.99 & 9.59 & 19.12 \\

GaussianFormer-2$^{\dagger}$~\cite{huang2025gaussianformer2}
& 30.56 & 20.02
& \underline{20.15} & \textbf{12.99} & 27.61 & 30.23
& 11.19 & \underline{15.31}
& 12.64 & 9.63 & 13.31 & \underline{22.26}
& 39.68 & 23.47 & 25.62 & 23.20 & 12.25 & 20.73 \\

SurroundOcc~\cite{wei2023surroundocc}
& 31.49 & 20.30
& \textbf{20.59} & 11.68
& \underline{28.06} & \underline{30.86}
& 10.70 & 15.14
& \textbf{14.09} & \textbf{12.06}
& \underline{14.38} & \underline{22.26}
& 37.29 & \underline{23.70} & 24.49 & 22.77 & 14.89 & 21.86 \\

GPOcc++~\cite{zhou2026gpoccplus}
& \underline{33.28} & \underline{20.78}
& 19.72 & 11.59 & 26.36 & 28.31 & \textbf{12.69} & 14.76
& 12.22 & \underline{10.67} & 13.62 & 20.04
& \underline{42.07} & \textbf{26.93} & \textbf{27.30}
& \underline{26.34} & \underline{15.27} & \underline{24.61} \\

\midrule
\rowcolor{softteal}
\textbf{CGOcc (Ours)}
& \textbf{34.65} & \textbf{21.95}
& 20.02 & \underline{12.83}
& \textbf{31.88} & \textbf{33.92}
& \underline{12.29} & \textbf{15.37}
& 11.98 & 10.17
& \textbf{15.90} & \textbf{24.62}
& \textbf{42.73} & 23.65
& \underline{25.63} & \textbf{26.58}
& \textbf{16.81} & \textbf{26.77} \\
\bottomrule
\end{tabular}%
}

\endgroup

\par\medskip

\begingroup
        \includegraphics[
            width=\textwidth,
            height=0.84\textheight,
            keepaspectratio
        ]{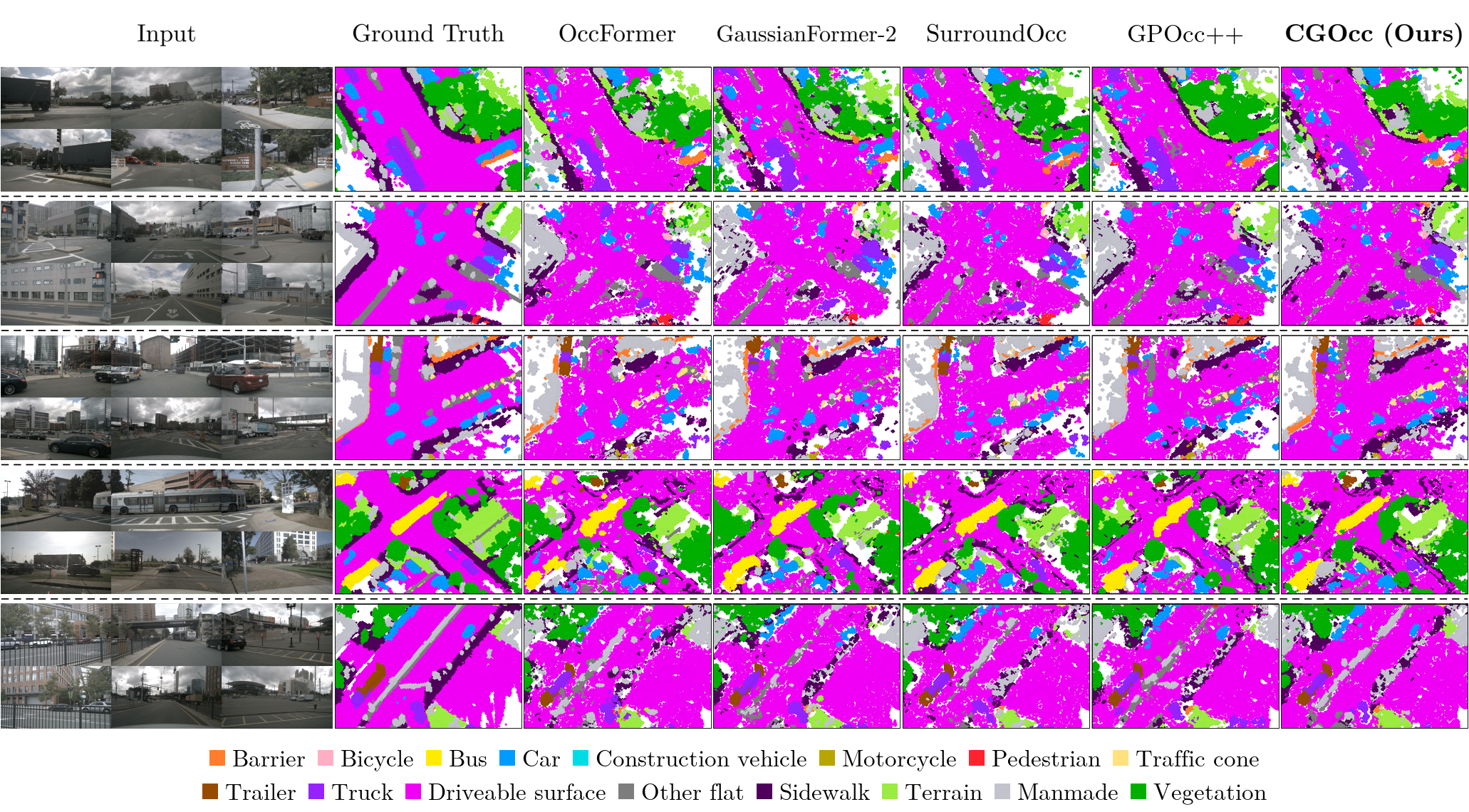}

    \captionof{figure}{
    \textbf{Qualitative comparison on SurroundOcc-nuScenes
    validation set.}
    }
    \label{fig:surroundocc_results}
\endgroup

\end{table*}

\end{document}